\documentclass[journal]{IEEEtran}
\usepackage{cite}

\ifCLASSINFOpdf
\else
\fi
\usepackage{amsmath}
\usepackage{amsfonts}
\usepackage{graphicx}
\usepackage{float}
\usepackage{booktabs}
\usepackage{multirow}
\usepackage[table,xcdraw]{xcolor}
\usepackage{overpic}
\usepackage{lineno}  
\begin{document}
%
\title{Self-supervised Depth Estimation Leveraging Global Perception and Geometric Smoothness \\ Using On-board Videos}
%
%
%

\author{Shaocheng~JIA,~\IEEEmembership{}
        Xin~PEI,~\IEEEmembership{}
        Wei~YAO,~\IEEEmembership{}
        and S.C.~WONG~\IEEEmembership{}
\thanks{This work has been submitted to the IEEE for possible publication. Copyright may be transferred without notice, after which this version may no longer be accessible. The work described in this paper was supported by the following: a collaborative project between China and Sweden on research, development, and innovation within life sciences and road traffic safety (Grant No. 2018YFE0102800); the National Natural Science Foundation of China (Grant No. 71671100); the Guangdong - Hong Kong - Macau Joint Laboratory Program of the 2020 Guangdong New Innovative Strategic Research Fund, Guangdong Science and Technology Department (Project No. 2020B1212030009); a grant from the Research Grants Council of the Hong Kong, Special Administrative Region, China (Project No. PolyU 25211819); and a grant from the Hong Kong Polytechnic University (Project No. G-YBZ9). The fourth author was supported by the Francis S Y Bong Professorship in Engineering. (Corresponding author: Wei YAO)}
\thanks{Shaocheng JIA is with the Department of Land Surveying and Geo-Informatics, The Hong Kong Polytechnic University, Hong Kong, China, and the Department of Automation, Tsinghua University, Beijing, China; Xin PEI is with the Department of Automation, Beijing National Research Center for Information Science and Technology, Tsinghua University, Beijing, China; Wei YAO is with the Department of Land Surveying and Geo-Informatics, The Hong Kong Polytechnic University, Hong Kong, China; and S.C. WONG is with the Department of Civil Engineering, The University of Hong Kong, and Guangdong - Hong Kong - Macau Joint Laboratory for Smart Cities, China.}
}
\maketitle

\begin{abstract}
Self-supervised depth estimation has drawn much attention in recent years as it does not require labeled data but image sequences. Moreover, it can be conveniently used in various applications, such as autonomous driving, robotics, realistic navigation, and smart cities. However, extracting global contextual information from images and predicting a geometrically natural depth map remain challenging. In this paper, we present DLNet for pixel-wise depth estimation, which simultaneously extracts global and local features with the aid of our depth Linformer block. This block consists of the Linformer and innovative soft split multi-layer perceptron blocks. Moreover, a three-dimensional geometry smoothness loss is proposed to predict a geometrically natural depth map by imposing the second-order smoothness constraint on the predicted three-dimensional point clouds, thereby realizing improved performance as a byproduct. Finally, we explore the multi-scale prediction strategy and propose the maximum margin dual-scale prediction strategy for further performance improvement. In experiments on the KITTI and Make3D benchmarks, the proposed DLNet achieves performance competitive to those of the state-of-the-art methods, reducing time and space complexities by more than $62\%$ and $56\%$, respectively. Extensive testing on various real-world situations further demonstrates the strong practicality and generalization capability of the proposed model.
\end{abstract}. 
    

%
\IEEEpeerreviewmaketitle

\section{Introduction}
%
%
%
%
\begin{figure}[htbp]
    \centering
    \begin{overpic}[scale=0.091]{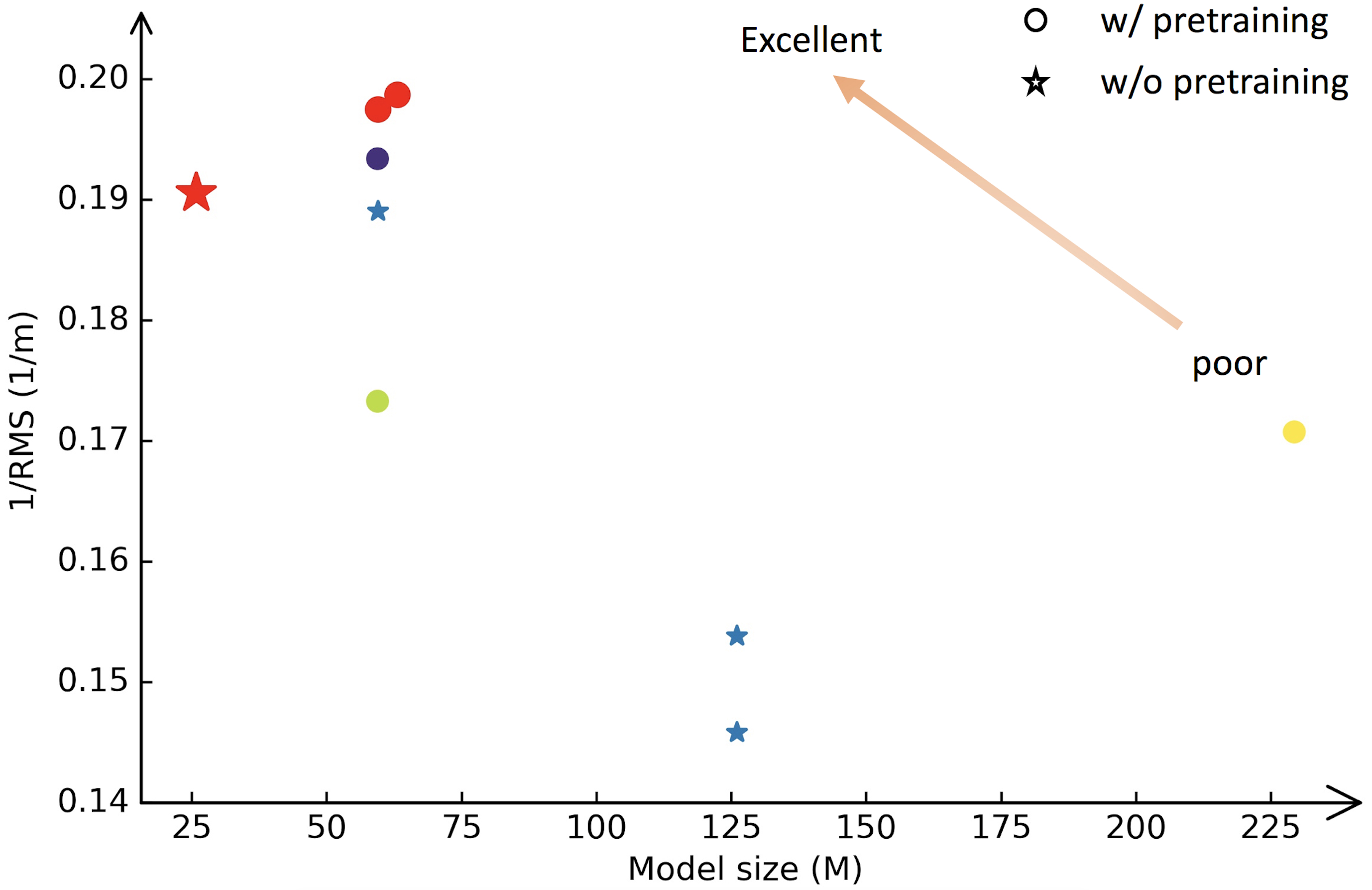}
    \put(11,53){Our(LL)}
    \put(16,60){Our(CC)}
    \put(32,57){Our(CL)}
    \put(30,52){Monodepth2 \cite{godard2019digging}}
    \put(20,45){Monodepth2 \cite{godard2019digging}}
    \put(15,30){SC-SfMLearner \cite{bian2019unsupervised}}
    \put(57,17){Yang et al. \cite{yang2017unsupervised}}
    \put(57,10){SfMLearner \cite{zhou2017unsupervised}}
    \put(70,28){GeoNet-Resnet \cite{yin2018geonet}}
    \end{overpic}
    \caption{Performance of self-supervised depth estimation and other state-of-the-art methods on the KITTI dataset \cite{geiger2013vision}. CC, CL, and LL represent different network architectures, in which C and L represent convolutional neural networks (CNNs) and the proposed encoder/decoder. For example, CL indicates an encoder and decoder adopting CNNs and the proposed decoder, respectively.}
    \label{performance}
\end{figure}
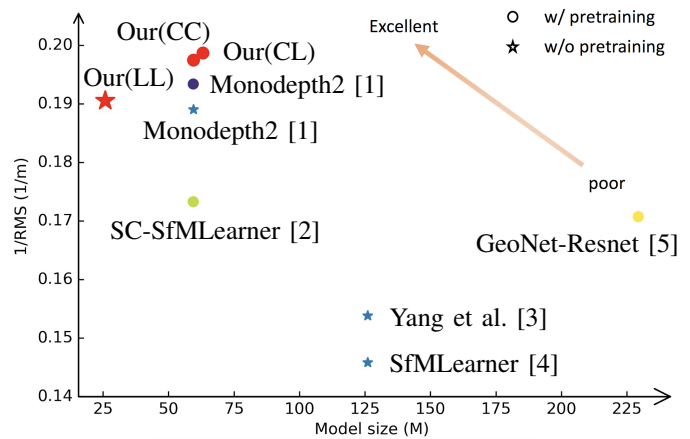

\IEEEPARstart{R}{ecovering} a scene’s depth information plays a significant role in three-dimensional (3D) reconstruction, robot navigation, and scene understanding. The depth information of a scene can be obtained using two types of sensors: active detection (e.g., light detection and ranging (LiDAR) sensors) and passive receiving (e.g., camera sensors). Using LiDAR, 3D point cloud data can be directly obtained by scanning a scene; this method is accurate but is expensive for routine use. Alternatively, image data from a camera sensor can be used to recover the 3D information.

Specifically, a stereo vision system can follow the epipolar geometry restrictions to recover the depth information in a straightforward manner, but this approach necessitates a binocular camera. In most cases, however, monocular camera data is preferred considering the energy consumption and cost constraints. Therefore, monocular depth estimation, as a convenient and economical method of recovering depth information, has attracted the attention of many scholars from diverse research fields. Unfortunately, extracting 3D information from a monocular vision system is challenging because of its inherent ill-conditioning.

Current monocular depth estimation methods can be divided into supervised and self-supervised methods depending on whether the ground truth is used during training. In supervised depth estimation, the ground truth depth map is used to train a deep neural network (DNN), which directly fits the relationship between the RGB image and the depth map and imposes some priors by designing different loss functions as well as devising several network variants to better extract the features. In self-supervised depth estimation, first proposed by Zhou et al. \cite{zhou2017unsupervised}, warping-based view synthesis is used as supervision to train the depth and pose networks. This approach does not require any labeled data and can simultaneously recover the depth and movement information. Although the depth map is relative, the absolute depth can be easily obtained with the aid of other information, such as real velocity from the global positioning system and the flat road assumption \cite{absolutedepthXue2020}.

In the past few years, both supervised and self-supervised depth estimation approaches have taken advantage of the powerful feature extracting ability of convolutional neural networks (CNNs). However, capturing global contextual information using pure CNNs is difficult because of the limited kernel size. To overcome this drawback, numerous studies have applied conditional random fields (CRFs) and Markov random fields (MRFs) \cite{cao2017estimating,eigen2015predicting,li2015depth,liu2015deep,mousavian2016joint,xu2017multi,xu2018structured,karschdepth,saxena20083}. Nevertheless, CRFs and MRFs are difficult to optimize, as is applying them to build an end-to-end model.

Predicting a geometric smoothness depth map facilitates both quantitative and qualitative evaluations. However, the smoothness loss measures used in previous works solely apply constraints to the two-dimensional (2D) depth map, and the 3D geometry properties of the scene are not considered. In addition, the multi-scale prediction strategy is often applied to overcome the gradient locality issue. Moreover, previous works have only used four-scale prediction frameworks, which raises the learning difficulty of the networks and thus negatively affects the performance.

To tackle the aforementioned issues, applying Linformer we propose a depth Linformer network (DLNet), a full-Linformer-based model, to concurrently capture global and local features, thereby improving the performance of self-supervised depth estimation \cite{Linformer}. Although many trials of applying the Transformer model \cite{Transformer} to computer vision tasks have been reported, to the best of our knowledge, few have applied pure Transformer or Linformer networks to perform pixel-wise tasks. Instead, researchers have attempted to either extract features (encoding) or predict results (decoding) using CNNs. To the best of our knowledge, the present study is the first to perform pixel-wise depth estimation with a full-Linformer-based model. Moreover, to further improve the quantitative and qualitative results, we explore geometry properties and multi-scale prediction.

Our contributions can be summarized as follows:
\begin{itemize}

\item To effectively extract global and local features, we propose a soft-split multi-layer perceptron (SSMLP) block and a depth Linformer block (DLBlock) to build the DLNet, the depth decoder, and the pose decoder.

\item We propose a 3D geometry smoothness (3DGS) loss to obtain a natural and geometry-preserving depth map by applying second-order smoothness constraints on the 3D point clouds rather than on the 2D depth map.

\item We present a maximum margin dual-scale prediction (MMDSP) strategy to overcome the gradient locality issue while concurrently saving computational resources and boosting performance.

\item Compared with state-of-the-art methods, the proposed model achieves competitive performance on the KITTI \cite{geiger2013vision} and Make3D \cite{saxena2008make3d} benchmarks but with a lightweight configuration and without pre-training. Furthermore, the promising qualitative results on the Cityscapes dataset \cite{cordts2016cityscapes} and real-world scenarios demonstrate the proposed model’s strong generalization capability and practicality.

\end{itemize}

The remainder of this paper is organized as follows. Section II introduces related works. Section III mathematically defines the problem and presents notational conventions. Section IV presents the model design and loss functions. Section V reports on detailed experiments, and Section VI discusses the limitations of the proposed model. Section VII draws the conclusions.

 
\section{RELATED WORK} 
In this section, we review the literature related to supervised depth estimation, self-supervised depth estimation, and transformer networks for computer vision.

\subsection{Supervised depth estimation}
Prior to advances in deep learning algorithms, monocular depth estimation was largely obtained by devising efficient handcrafted features to capture the 3D information \cite{karschdepth,saxena20083}. For example, Saxena et al. \cite{saxena20083} extracted absolute and relative depth features from the textures and statistical histograms of images, respectively, and integrated the extracted features and MRFs to predict the final depth map. Research on depth estimation has since proliferated, mainly focusing on exploring monocular cues in images \cite{baig2016coupled,choi2015depth,furukawa2017depth,zoran2015learning}.

However, obtaining abstract and deep features through such manual design is challenging. Fortunately, CNNs can aid in extracting abstract and complicated features from images. To the best of our knowledge, Eigen et. al. \cite{eigen2015predicting} were the first to apply CNNs for monocular depth estimation, and numerous variants, focusing on network structure design, have been proposed since \cite{chen2016single,eigen2014depth,eigen2015predicting,laina2016deeper,li2017two}. In addition, to overcome the spatial locality of the convolution operator, CRFs and recurrent neural networks (RNNs) have been introduced to capture the global information of an image \cite{cao2017estimating,eigen2015predicting,li2015depth,liu2015deep,mousavian2016joint,xu2017multi,xu2018structured, almalioglu2019ganvo, cs2018depthnet, grigorev2017depth, mancini2017toward, tananaev2018temporally, wang2019recurrent, mypaper2020}.

Typically, depth estimation is regarded as a pixel-wise regression problem, but it can also be cast as a classification problem by discretizing the continuous depth into many intervals so as to predict a specific label for each pixel \cite{cao2017estimating, fu2018deep}.

\begin{figure*}[htbp]
    \centering
    \includegraphics[width=\textwidth]{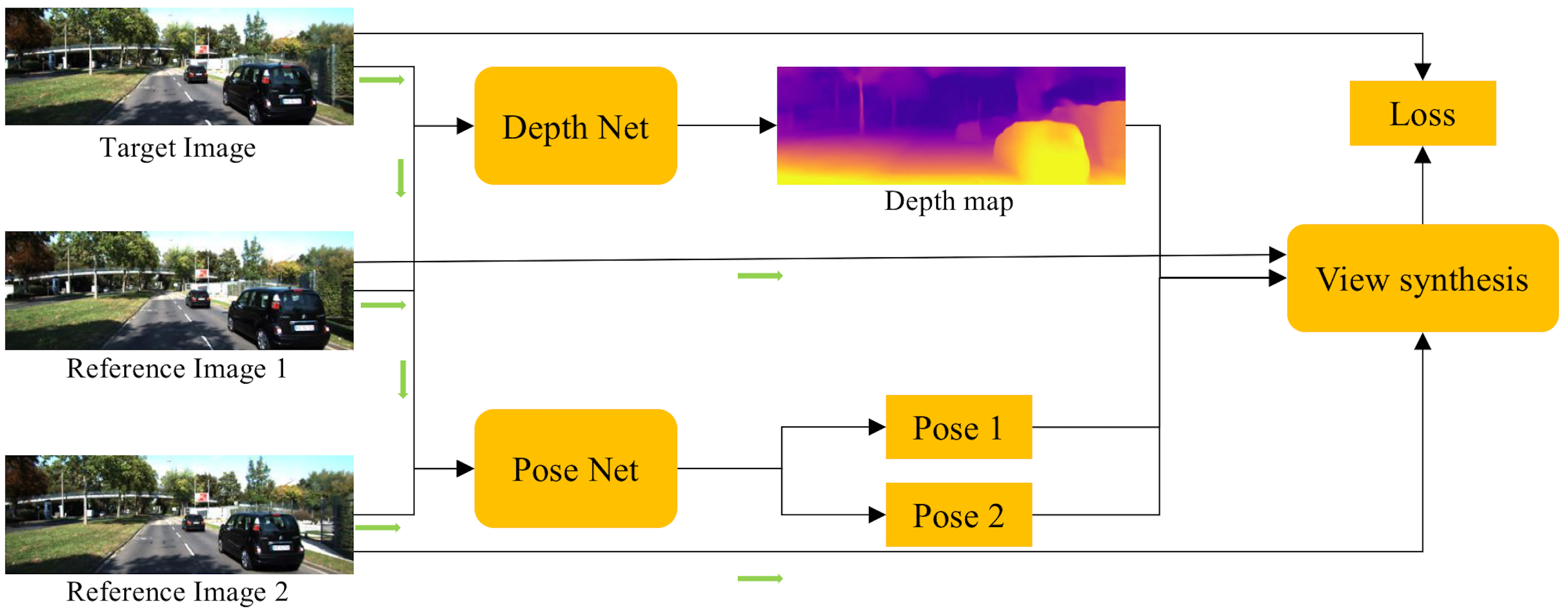}
    \caption{Self-supervised monocular depth estimation system. Pose $i$ is the predicted ego-motion vector between the target image and reference image $i$ ($i = 1, 2$).}
    \label{framework}
\end{figure*}

\subsection{Self-supervised depth estimation}

Differing from supervised depth estimation, self-supervised depth estimation uses warping-based view synthesis to reconstruct the target image and then trains the model by computing the difference between the reconstructed and target images \cite{zhou2017unsupervised}. Self-supervised depth estimation relies on monocular data and does not require any labeled data, an advantage that has attracted many researchers \cite{chen2019towards, garg2016unsupervised, ranjan2019competitive, yin2018geonet, zhan2018unsupervised, zhou2019unsupervised, zhou2017unsupervised, godard2017unsupervised, kuznietsov2017semi, almalioglu2019ganvo, feng2019sganvo}.

However, recovering a scene’s structure from motion (SfM) is inherently problematic for some special cases, such as moving objects and occlusions. To bridge these application gaps, a series of outstanding studies have been conducted. Representative studies include the following: Godard et al. \cite{godard2019digging} proposed a minimal reprojection loss function to effectively improve the occlusion/disocclusion problem; Casser et al. \cite{casser2019depth} used the advanced semantic segmentation model to mask potential moving objects out, thus excluding their influence; Zhou et al. \cite{zhou2017unsupervised} proposed multi-scale training for solving the gradient locality issue caused by low textures; Bian et al. \cite{bian2019unsupervised} presented a geometry consistency loss function for achieving scale-consistent depth and ego-motion estimation within a continuous sequence; and Jia et al. \cite{mypaper2021} modeled the prediction uncertainty and relationships between depths to realize a reliable and practical depth estimation system.

Moreover, Park et al. \cite{park2019high} and Yang et al. \cite{yang2019fast} have integrated data from multiple sensors, such as LiDAR sensors, visual odometers, and cameras, for improving the inference efficiency and accuracy.

\subsection{Transformer for computer vision}

Transformer \cite{Transformer}, an attention-based model initially proposed for natural language processing (NLP), is efficient at capturing long-range dependencies between items. Numerous recent studies have applied Transformer to computer vision tasks by reshaping square images to sequence-like data, either using Transformer for CNN feature processing \cite{T4C1, T4C2, T4C3, T4C4} or for feature extraction \cite{C4T1, C4T2, C4T3, C4T4}. Specifically, in the former application, Transformer is used as a decoder for predictions, whereas in the latter, Transformer substitutes CNNs and is used as an encoder for feature extraction.

However, applying the classic Transformer to pixel-wise computer vision tasks, such as semantic segmentation and depth estimation, is difficult because it requires large storage and computational resources for processing long sequence data. Hence, for efficiency in pixel-wise tasks, CNNs are generally used at the beginning or end of the network \cite{C4T3,T4C4}.

In addition, when rigidly splitting an image into many patches and using them as the input of the network, it is difficult to capture delicate features, such as edges \cite{C4T4}. Consequently, Yuan et al. \cite{C4T4} proposed a tokens-to-token strategy to aggregate neighboring features.

In summary, the literature reveals that effectively and efficiently extracting global and local information from images remains challenging, especially when using the emerging Transformer model. Moreover, no studies have examined the second-order geometric smoothness of the predicted point clouds. These research gaps have inspired the present work.

\section{PROBLEM SETUP}
A self-supervised monocular depth estimation system comprises two parts, namely the depth estimation network and the pose estimation network, denoted as $\Psi_D$ and $\Psi_E$, respectively. Given a continuous image sequence $I=(I_r^{-k},..., I_r^{-1}, I_t, I_r^1,..., I_r^k), I_t, I_r^i \in {\mathbb{R}^{h\times w \times c}}$, the depth estimation network $\Psi_D$ solely takes the target image $I_t$ as the input to predict its depth map; this can be mathematically defined as: $\Psi_D: I \in {\mathbb{R}^{h\times w \times c}} \to D \in \mathbb{R}^{h \times w}$, where $D$ is the predicted depth map of the input image $I$. Differing from the depth estimation network, the pose estimation network takes the whole sequence as the input and predicts the ego-motion for each image pair $(I_t, I_r^i), i=-k,...,-1, 1,...k$; this can be mathematically presented as: $\Psi_E: I \in {\mathbb{R}^{(2k+1) \times h\times w \times 3}} \to T \in \mathbb{R}^{2k \times 6}$, where $T$ is a pose matrix of describing the movement between the target and reference images.

Theoretically, given the depth map of the target image and the ego-motion between the target and reference images, the target image can be reconstructed from the reference images by warping-based view synthesis \cite{zhou2017unsupervised}, which can be mathematically defined as Eq. \ref{proj}; here, $p_{r}$, $K$, $E_{t \to r}$, $D_{t}(p_{t})$, $K^{-1}$, and $p_{t}$ represent the coordinate in the reference image, the camera intrinsic matrix ($\mathbb{R}^{3 \times 3}$), the transform matrix between the target image and reference images, the depth corresponding to $p_{t}$, the inverse matrix of $K$, and the coordinate in the target image, respectively.

\begin{equation}
\label{proj}
  p_r\sim KE_{t\rightarrow r}D_t(p_t)K^{-1}p_t.
\end{equation}

During the training phase, the reconstructed loss is computed with respect to the difference of the target and reconstructed images to train the system. Notably, the depth and pose estimation networks are trained cooperatively, but they can work separately during the testing phase.

\section{METHOD}
In this section, we first illustrate the entire monocular depth estimation system. Subsequently, the proposed DLNet is introduced. Thereafter, the DLNet-based depth and pose estimation networks are presented. Finally, the loss functions used in this paper are presented.

\begin{figure*}[htbp]
    \centering
    \includegraphics[width=\textwidth]{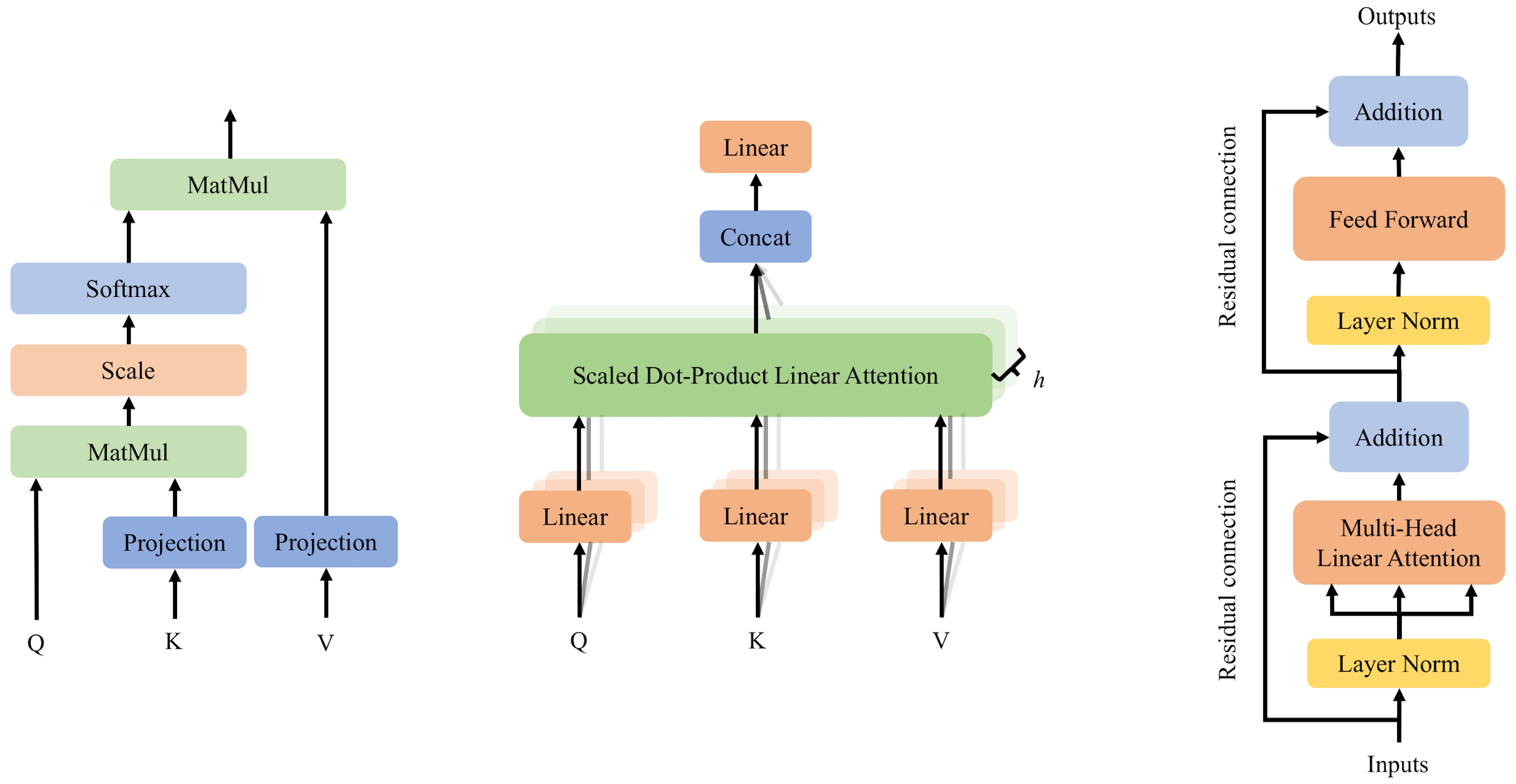}
    \caption{Architectures of the Linformer layer and its components. Left to right: scaled dot-product linear attention (SDPLA), multi-head linear attention (MHLA), and Linformer block.}
    \label{Linformer}
\end{figure*}

\subsection{Model overview}
The self-supervised monocular depth estimation system concurrently performs depth and pose estimation during training, as shown in Fig. \ref{framework}. Following Zhou et al. \cite{zhou2017unsupervised}, the length of the image sequence used for training is set to $3$, and the middle frame of the sequence is regarded as the target image that requires depth estimation using depth net. In contrast, the whole sequence of the three frames is used for pose estimation.

After obtaining the depth map and pose vectors, warping-based view synthesis can be performed for reconstructing the target image from the reference images. Then, the reconstruction loss, namely the difference between the target and reconstructed images, is computed to train the system. In the following subsections, the proposed DLNet, depth net, pose net, and loss function are introduced.

\subsection{Depth Linformer Network (DLNet)}
\subsubsection{Linformer} Transformer uses the scaled dot-product attention (SDPA) mechanism to perform feature aggregation, which can be intuitively described as mapping a query and a set of key–value pairs to an output \cite{Transformer}. In particular, the query, keys, values and output are all vectors with dimensions $d_k$, $d_k$, $d_v$, and $d_v$, respectively. However, in practice, we pack a set of queries together to perform the attention computation simultaneously.

For clarity, we denote the queries of a sequence of length $n$ as the matrix $Q \in \mathbb{R}^{n \times d_k}$. Accordingly, the keys and values are denoted as the matrices $K \in \mathbb{R}^{n \times d_k}$ and $V \in \mathbb{R}^{n \times d_v}$. Thus, the SDPA can be defined as Eq. \ref{classic_attention}:
\begin{equation}
\label{classic_attention}
  Attention(Q,K,V)=\begin{matrix} \underbrace{softmax(\frac{QK^T}{ \sqrt{d_k}})}V \\ P: n \times n \end{matrix}.
\end{equation}

The attention matrix $P \in \mathbb{R}^{n \times n}$ is obtained by multiplying two $n \times d_k$ matrices, which requires $O(n^2)$ time and space complexities with respect to the length of the sequence. In many cases, the sequence length requires a prohibitively large amount of storage and computational resources when using the Transformer model, especially for pixel-wise computer vision tasks.

To overcome this problem, Wang et al. \cite{Linformer} proposed a linear complexity ($O(n)$) Transformer based on the low-rank property of the attention matrix $P$, called Linformer, significantly reducing the time and space complexities. Specifically, two learnable
matrices $E, F \in \mathbb{R}^{k \times n}$ are used to project the original $n \times d$-dimensional matrices $K$ and $V$ into $k \times d$-dimensional ($k << n$) space; accordingly, SDPA can be rewritten as scaled dot-product linear attention (SDPLA) (Eq. \ref{linear_attention}). For simplicity, we do not differentiate between $d_k$ and $d_v$ in the following text.
\begin{equation}
  \label{linear_attention}
  \overline{Attention}(Q,K,V)=
  \begin{matrix} \underbrace{softmax(\frac{Q(EK)^T}{\sqrt{d_k}})}(FV) \\ 
  \overline{P}: n \times k 
  \end{matrix}.
\end{equation}

Accordingly, the multi-head linear attention (MHLA) can be described as follows (Eq. \ref{multi_linear_attention}):
\begin{equation}
\label{multi_linear_attention}
\begin{split}
\begin{aligned}
  MHLA(Q,K,V)&=Concat(head_1, head_2,..., head_h)W^O \\ s.t. \quad head_i &= \overline{Attention}(QW_i^Q,KW_i^K,VW_i^V) \\ i&=1,...,h
\end{aligned}
\end{split},
\end{equation}
where $W^O \in \mathbb{R}^{hd \times d}$ and $W_i^Q, W_i^K, W_i^V \in \mathbb{R}^{d \times d}$ are learnable parameters, and $h$ is the number of heads. Finally, the Linformer block is designed by integrating MHLA and multi-layer perceptron (MLP); the detailed structures are shown in Fig. \ref{Linformer}.

\subsubsection{Depth Linformer block (DLBlock)} Most studies have rigidly divided the image into many patches, flattening the patches to vectors and using them as the input of the Transformer model. However, in this approach, obtaining fine features of the image, such as edges, is challenging because of the lack of communication between the patches. Furthermore, the original Transformer \cite{Transformer} and Linformer \cite{Linformer} models cannot dynamically change the feature map resolution, resulting in high computational and storage costs, especially for image processing.

\begin{figure}[htbp]
    \centering
    \includegraphics[scale=0.098]{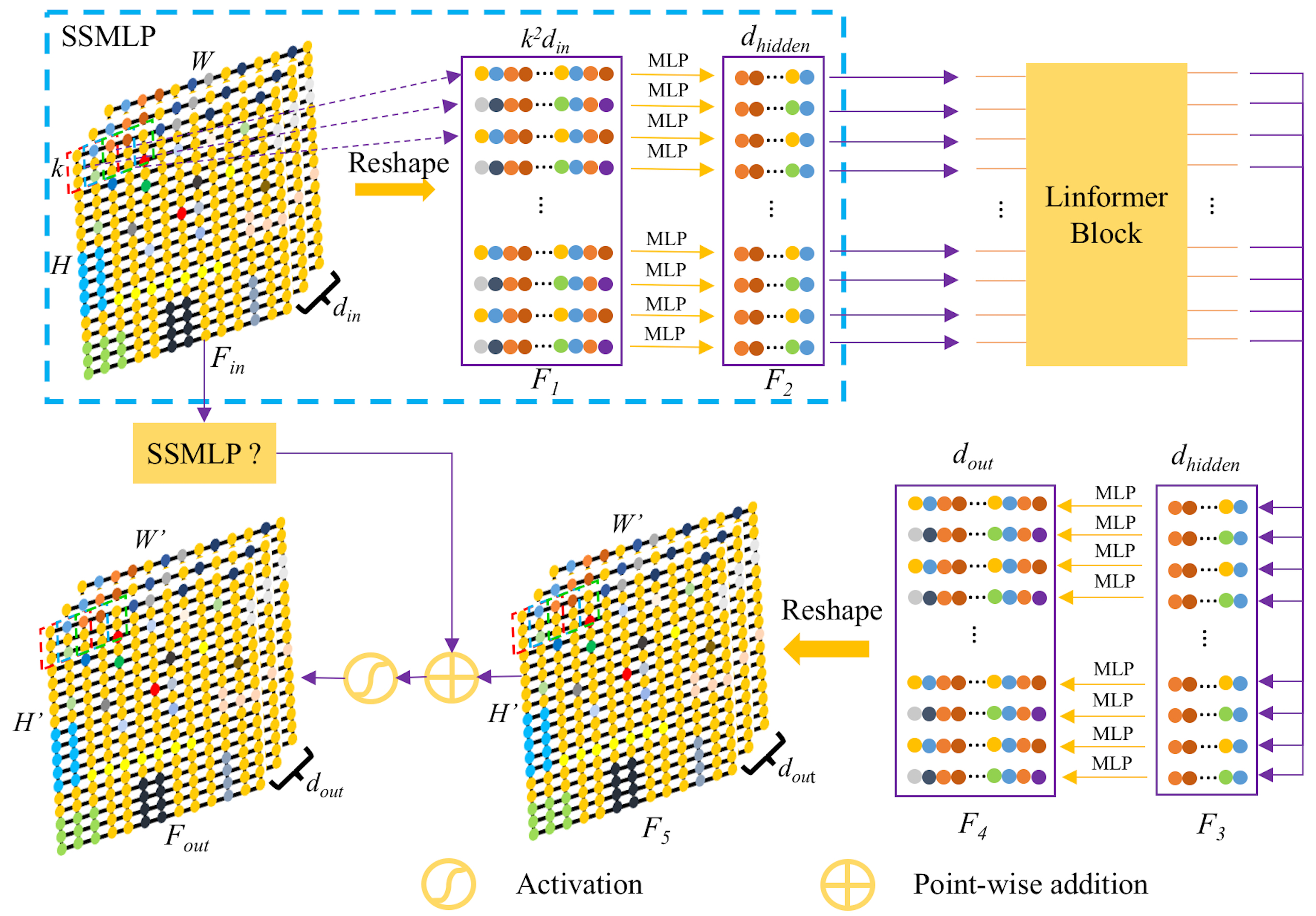}
    \caption{Architecture of the proposed Depth Linformer block (DLBlock).}
    \label{DLBlock}
\end{figure}

To overcome this, we introduce the soft-split multi-layer perceptron (SSMLP) block to promote communication between the patches, thereby simultaneously adjusting the feature map size and reducing the computational and storage costs. Subsequently, the features obtained from the SSMLP are delivered to the Linformer block for extracting the global features. Figure \ref{DLBlock} illustrates the detailed structure of the proposed DLBlock.

\begin{figure*}[htbp]
    \centering
    \includegraphics[width=\textwidth]{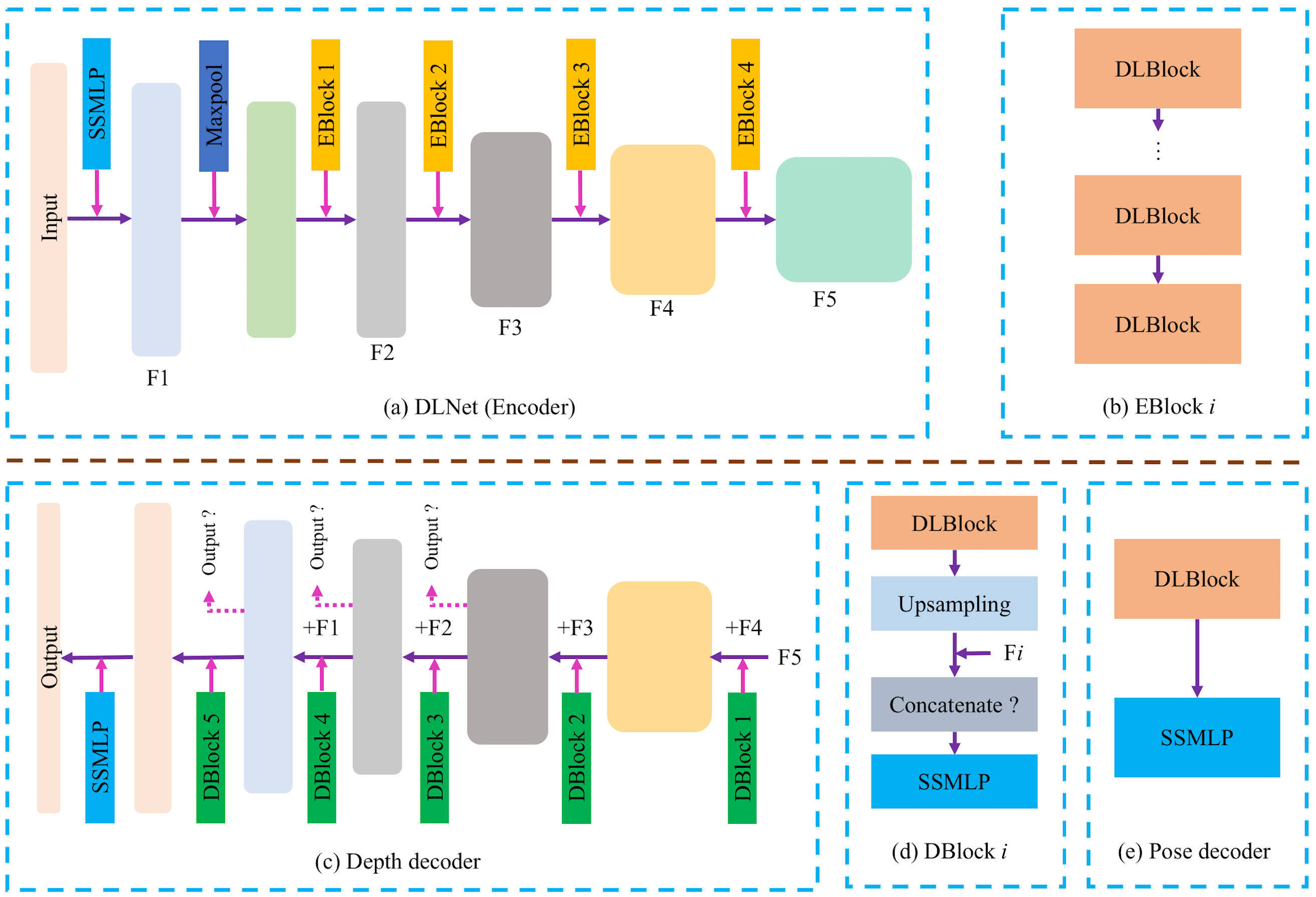}
    \caption{Architectures of the proposed depth Linformer network (DLNet), depth decoder, and pose decoder. EBlock $i$ and DBlock $i$ represent encoder block $i$ and decoder block $i$, respectively. +F$i$ indicates taking in the F$i$ feature via the corresponding shortcut connection. In DBlocks, the concatenation operation is used, if F$i$ is available. Similarly, multi-scale outputs are available for the different configurations.}
    \label{networks}
\end{figure*}

Let us denote the input feature as $F_{in} \in \mathbb{R}^{H \times W \times d_{in}}$, where $H$, $W$, and $d_{in}$ represent the height, width and dimension number of the input feature, respectively. For aggregating the local feature, a moving window of size $k \times k$, stride $s$, and padding $l$, is used to reshape the input feature to the sequence data $F_1 \in \mathbb{R}^{(H^{'}W^{'}) \times (kkd_{in})}$ in a soft-split manner, wherein $H^{'}$ and $W^{'}$ are derived from Eq. \ref{height_width}:
\begin{equation}
\label{height_width}
\begin{aligned}
    H^{'} &= \lfloor \frac{H - k + 2l}{s} + 1 \rfloor\\
    W^{'} &= \lfloor \frac{W - k + 2l}{s} + 1 \rfloor
\end{aligned},
\end{equation}
where $\lfloor \cdot \rfloor$ denotes rounding down.

When the stride $s$ and moving window size $k$ satisfy $s \leq k - 1$, adequate overlapping exists (soft split) for capturing the fine details of the image. However, in this case, the dimension of the feature is multiplied by $k^{2}$, significantly raising the computational and storage costs. Accordingly, MLP is used for dimension reduction (Eq. \ref{ssmlp1}):
\begin{equation}
\label{ssmlp1}
    F_2 = Act(LN(F_1)W_{MLP}^{1} + b_{1}),
\end{equation}
where $F_2 \in \mathbb{R}^{(H^{'}W^{'}) \times d_{hidden}}$ represents the transformed low-dimensional features; $W_{MLP}^{1} \in \mathbb{R}^{kkd_{in} \times d_{hidden}} $ and $b_{1} \in \mathbb{R}^{d_{hidden}}$ are learnable parameters; and $d_{hidden}$, $LN(\cdot)$, and $Act(\cdot)$ represents the target dimension of dimension reduction, layer normalization, and activation function, respectively. The broadcasting mechanism is automatically performed for the foregoing additive operation. Furthermore, $d_{hidden}$ is given by Eq. \ref{hiddendimension}:
\begin{equation}
\label{hiddendimension}
    d_{hidden} = \frac{min(d_{in}, d_{out})}{c},
\end{equation}
where $c$ is a hyperparameter and $d_{out}$ is the output dimension.

The aforementioned transformations and computations, called SSMLP, effectively conduct local feature extracting, with the aid of the soft split and the MLP layer.

Subsequently, the feature $F_2$ goes through the Linformer block to capture the global information, thereby obtaining the feature $F_3 \in \mathbb{R}^{(H^{'}W^{'}) \times d_{hidden}}$. Then, another MLP layer performs the dimension increase to change the feature $F_3$ to $F_4 \in \mathbb{R}^{(H^{'}W^{'}) \times d_{out}}$; this is followed by the reshaping of $F_4$ to $F_5 \in \mathbb{R}^{H^{'} \times W^{'} \times d_{out}}$, which can be formulated as Eq. \ref{ssmlp2}:
\begin{equation}
\label{ssmlp2}
    F_5 = Reshape(LN(F_3)W_{MLP}^{2} + b_{2}).
\end{equation}

Finally, we bring in the initial feature $F_{in}$ through a residual connection, which is stated in Eq. \ref{shortcut}:
\begin{equation}
\label{shortcut}
    F_{out} = 
    \begin{cases}
    Act(F_5 + SSMLP(F_{in})) &size(F_{in}) \neq size(F_{5}) \\
    Act(F_5 + F_{in}) &size(F_{in}) = size(F_{5})
    \end{cases},  
\end{equation}
where $F_{out} \in \mathbb{R}^{H^{'} \times W^{'} \times d_{out}}$ is the output feature.

In summary, our DLBlock mainly comprises the following three components:
\begin{itemize}
\item \textbf{SSMLP}. SSMLP is critical for extracting the local feature and improving the efficiency;

\item \textbf{Linformer block.} The Linformer block is crucial for capturing global features, automatically shifting the focus to the more important features through the inner self-attention mechanism;

\item \textbf{Residual connection.} Residual connection, a proven technique, can improve the gradient explosion and network degradation, concurrently avoiding to the extent possible the information loss caused by changes in the feature map size and dimensions.
\end{itemize}
In what follows, DLNet, a DLBlock-based network, is introduced.

\subsubsection{Depth Linformer Network (DLNet)} Inspired by the success of the CNNs, a pyramid-like structure of gradually reducing the feature map size is adopted when devising the DLNet. As shown in Fig. \ref{networks} (a), an SSMLP layer is first used to embed the input image and simultaneously decrease the feature map's resolution. After a MaxPooling layer, the four-stage feature transformations are performed via the encoder blocks (Fig. \ref{networks} (b)), which includes the proposed DLBlock.

In the next subsection, the depth and pose networks are devised based on the proposed DLNet.

\subsection{Depth and pose estimation networks}
The proposed DLNet is considered the encoder in both the depth and pose estimation networks. The integral depth and pose estimation networks are presented in this subsection by further devising the decoders using the proposed components.

Figure \ref{networks} (c) illustrates the structure of the proposed depth decoder, which consists of a few decoder blocks and output heads. For each decoder block illustrated in Fig. \ref{networks} (d), a DLBlock is primarily used to perform the feature transformation, following which an upsampling layer is used to increase the resolution. Subsequently, the upsampled feature is stacked with the feature from the skip connection with respect to the channel, when the skip connection is available. Finally, a lightweight SSMLP layer is used for feature compression. Because of the availability of the skip connections, the residual connection in DLBlock is discarded to save computational and storage resources. For each output head, an SSMLP is used to predict the disparity map. 

Figure \ref{networks} (e) illustrates the structure of the proposed pose decoder, which simply consists of a DLBlock without the residual connection and an SSMLP layer.

Then, the depth and pose estimation networks can be obtained by simply integrating the proposed DLNet and the corresponding decoder.

\subsection{Losses}
In this subsection, a set of loss functions used for training the networks are presented. Specifically, basic losses that have been successfully applied in previous works are introduced. Subsequently, a novel 3D geometry smoothness (3DGS) loss function and the maximum margin dual-scale prediction (MMDSP) are presented. Thereafter, the final loss is shown.

\subsubsection{Basic losses} A strong assumption, the Lambertian reflection \cite{basri2003lambertian}, is imposed on all surfaces of the image, which makes the photometric constancy loss between the target image $I_{t}$ and the reconstructed image $I_{r}^{i}$ possible. Taking the robustness into account, we, therefore, choose the L1 norm to compute the photometric loss, which can be stated as Eq. \ref{photoloss}:
\begin{equation}
  \label{photoloss}
  \begin{aligned}
  L_p^{i} &= \| I_{i \to t}-I_t \|_1,\\
  s.t. \quad I_{i \to t} &= [proj(I_r^{i}, K, E_{t \to i})],\\
  i &=-k,...,-1,1,...,k,
  \end{aligned}
\end{equation}
where $I_{i \to t}$ represents the reconstructed image from the reference image $I_r^{i}$; $proj(\cdot)$ represents the reprojection function described in Eq. \ref{proj}; and $[\cdot]$ is the bilinear sampling operator, which is locally sub-differentiable. For simplicity, these notations are directly used in the rest of this paper.

However, the photometric loss is sensitive to illumination changes, particularly in complicated real-world scenarios. Consequently, following Godard et al. \cite{godard2019digging}, the structure similarity (SSIM) loss (Eq. \ref{ssim}) is used to improve this issue:
\begin{equation}
  \label{ssim}
  L_{ssim}^{i} = \frac{1 - SSIM(I_{i \to t}, I_t)}{2}.
\end{equation}

To address the problem of visual inconsistencies in the target and reference images, such as occlusion and disocclusion, we follow Godard et al. \cite{godard2019digging} in adopting the minimum reprojection loss (Eq. \ref{minproj}):
\begin{equation}
  \label{minproj}
  \begin{aligned}
  L_{min} &= min(L_{-k}, ..., L_{-1}, L_{1},...,L_{k}), \\
  s.t. \quad L_i &= \alpha L_{p}^{i} + (1-\alpha)L_{ssim}^{i},
  \end{aligned}
\end{equation}
where $\alpha$ is set as 0.15 following \cite{godard2019digging}.

Furthermore, we apply a simple binary mask proposed by Godard et al. \cite{godard2019digging} to avoid the influence of the static pixels caused by the static camera, an object moving at equivalent relative translation to the camera, and the low-texture regions, as follows (Eq. \ref{automasking}):
\begin{equation}
  \label{automasking}
  \mu_i = L_i < L_{io},
\end{equation}
where $\mu_i \in \left \{0, 1 \right \}$ is a binary mask and $L_{io}$ is the difference between the target image $I_t$ and the unwarped reference image $I_r^{i}$. Therefore, the reconstruction loss can be written as Eq. \ref{reconstructionloss}:
\begin{equation}
  \label{reconstructionloss}
  L_{recons} = min(\mu_{-k}L_{-k}, ..., \mu_{-1}L_{-1}, \mu_{1}L_{1},...,\mu_{k}L_{k}).
\end{equation}

Finally, the scalar reconstruction loss $\mathcal{L}_{recons}$ can be computed by averaging over each pixel and batch, as follows (Eq. \ref{finalreconstructionloss}):
\begin{equation}
  \label{finalreconstructionloss}
  \mathcal{L}_{recons} = \frac{1}{N_{b}} \sum_{b} \frac{1}{N_{p}} \sum_{p}L_{recons},
\end{equation}
where $N_{b}$ and $N_{p}$ are the batch size and the number of the pixels, and $b$ and $p$ represent the traversing of each sample and pixel.

\subsubsection{3D geometry smoothness (3DGS) loss} Generally, a smoothness loss is applied to obtain a smooth depth map. However, the smoothness loss used in previous works \cite{zhou2017unsupervised, godard2019digging} simply constrains the distance between the neighboring depths and does not take the geometric properties into account. Mathematically, the distance of the target depth from its neighbors can be directly minimized (Eq. \ref{naivesmooth}) to encourage a $\mathcal{C}^{0}$ smoothness depth map, which solely promotes continuity on depth values.

\begin{equation}
  \label{naivesmooth}
  min \quad \|D(p) - D(p_{neighbors}) \|.
\end{equation}

However, in this case, there are two major drawbacks as follows:
\begin{itemize}
\item \textbf{Non-differentiable artifacts.} The naive smoothness loss function does not consider the differentiability of the depth map, resulting in unnatural and non-differentiable artifacts, especially in the edge regions of the objects;
\item \textbf{Violation of the geometry structure.} The values from the close to the far regions in the depth map/disparity map increase/decrease monotonically with various granularities. Nevertheless, the naive smoothness loss applies identical weights over different positions, which breaks up the overall geometry structure of the scene.
\end{itemize}

Therefore, we propose the 3DGS, aimed at predicting a smooth, geometry-preserved, and natural depth map by imposing the gradual change constraint on the surface normals of the reconstructed 3D point clouds.

Primarily, we need to estimate the pixel-wise surface normal from the predicted depth map. Thus, the depth map is first reprojected to 3D space using Eq. \ref{pointproj}:
\begin{equation}
\label{pointproj}
  P \sim DK^{-1}p,
\end{equation}
where $p$, $K$, $D$, and $P$ represent the image coordinates, camera intrinsic matrix, depth map, and point clouds, respectively.
\begin{figure}[htbp]
    \centering
    \includegraphics[scale=0.1]{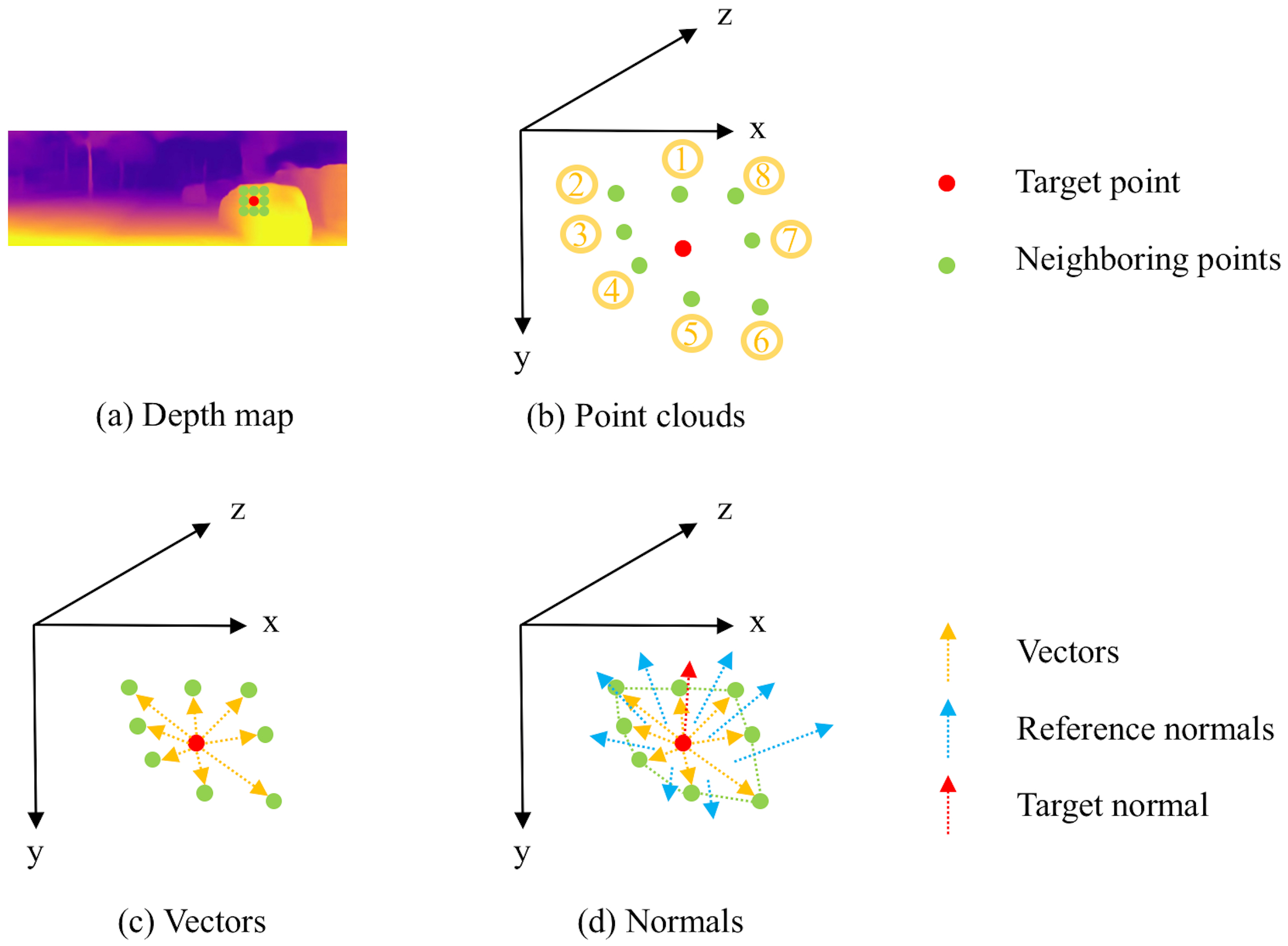}
    \caption{Illustration of the estimation of the surface normal.}
    \label{estimatingnormal}
\end{figure}

Then, the target point $P_t$ and its eight neighborhood points $P_i$ can be used to determine eight vectors $\overrightarrow{P_{t}P_{i}}$, where $i=1,2,...,8$. Any two arbitrary two neighboring vectors can determine a surface. For each surface, we can obtain the normal by computing the cross-product of the neighboring two vectors. Finally, the target surface normal is estimated by averaging over all reference normals, as shown in Eq. \ref{estimatingn}:
\begin{equation}
\label{estimatingn}
  \begin{aligned}
  \overrightarrow{n_{t}} &= \frac{1}{8} \sum_{i=1}^{8} \overrightarrow{n_i}, \\
  s.t. \quad \overrightarrow{n_i} &= \overrightarrow{P_{t}P_{i}} \otimes \overrightarrow{P_{t}P_{j}}, \\
  j &= \begin{cases}
  i + 1 & i + 1 \leq 8 \\
  1 & i + 1 > 8
  \end{cases},
   \end{aligned}
\end{equation}
where $\otimes$, $\overrightarrow{n_{t}}$, and $\overrightarrow{n_{i}}$ represent the cross-product operation, target normal, and reference normal, respectively.

Following the pixel-wise normal estimation, we apply the proposed 3DGS loss function, constraining and slowly changing the surface normals of the scene smoothly. First, consider a continuous space. Given a surface $\pi=f(x, y)$, which is defined on a two-dimensional space $\theta$ without any sharp points, the surface $\pi$ should be continuous if Eq. \ref{continuoussurface} holds:
\begin{equation}
\label{continuoussurface}
\lim_{\Delta x,\Delta y \rightarrow 0} f(x+\Delta x, y+\Delta y) - f(x,y) = 0, (x,y) \in \theta.
\end{equation}

In this case, the surface $\pi$ has $\mathcal{C}^{0}$ smoothness. If the surface normal is everywhere available, we can infer that the surface $\pi$ is first-order differentiable (please note that we assume that there are no sharp points in the surface $\pi$, such as the point $(0, 0)$ in the curve $f(x) = |x|$), which indicates that the surface $\pi$ has $\mathcal{C}^{1}$ smoothness. Finally, the gradual changes of the surface normal, namely the smooth surface normals, require the surface $\pi$ to be second-order differentiable, making the surface $\pi$ have $\mathcal{C}^{2}$ smoothness.

Therefore, we first define the distance between the two surface normals as the sine distance (Eq. \ref{normaldistance}) to achieve the surface normal smoothness:
\begin{equation}
\label{normaldistance}
d(\overrightarrow{n_{1}}, \overrightarrow{n_{2}}) = \S(\overrightarrow{n_{1}}, \overrightarrow{n_{2}}) = 1 - ( \frac{\overrightarrow{n_{1}} \cdot \overrightarrow{n_{2}}}{\|\overrightarrow{n_{1}}\|_{2}\|\overrightarrow{n_{2}}\|_{2}})^{2},
\end{equation}
where $\S$ is the sine distance operator. Thus, the proposed 3DGS loss can be described as Eq. \ref{geometrysmoothness}:
\begin{equation}
\label{geometrysmoothness}
\begin{aligned}
\mathcal{L}_{3DGS} = \frac{1}{N_b}\sum_{b}\frac{1}{N_p}&\sum_{p} \begin{matrix} \underbrace{ |e^{-\partial_{x}I}|\S_{x}A + |e^{-\partial_{y}I}|\S_{y}A} \\
\mathcal{C}^{1}, \mathcal{C}^{2} \; smoothness
\end{matrix} \\
& + \begin{matrix} \underbrace{ |e^{-\partial_{x}I}|\partial_{x}D^{-1} + |e^{-\partial_{y}I}|\partial_{y}D^{-1}} \\
\mathcal{C}^{0} \; smoothness
\end{matrix},
\end{aligned}
\end{equation}
where $\partial$, $A$ $D^{-1}$, and $I$ represent the gradient operator, estimated surface normal matrix, predicted disparity map, and color image, respectively. The exponential items slack the constraints on the edges for performing edge-aware prediction.

By requiring the 3DGS, the proposed model can predict a smooth and natural depth map, significantly improving the qualitative and quantitative performance, particularly in the edge regions.

\subsubsection{Maximum margin dual-scale prediction (MMDSP)} To overcome the gradient locality issue raised by the low-texture regions in the image, most previous works have used the multi-scale prediction strategy \cite{zhou2017unsupervised, godard2019digging}, as it is relatively easy to capture contextual information at a lower resolution so as to accurately predict the depth map for low-texture regions.

Accordingly, we adopt multi-scale training in our network. However, is the four-scale prediction used in previous works \cite{godard2019digging} necessary?
\begin{figure}[htbp]
    \centering
    \includegraphics[scale=0.13]{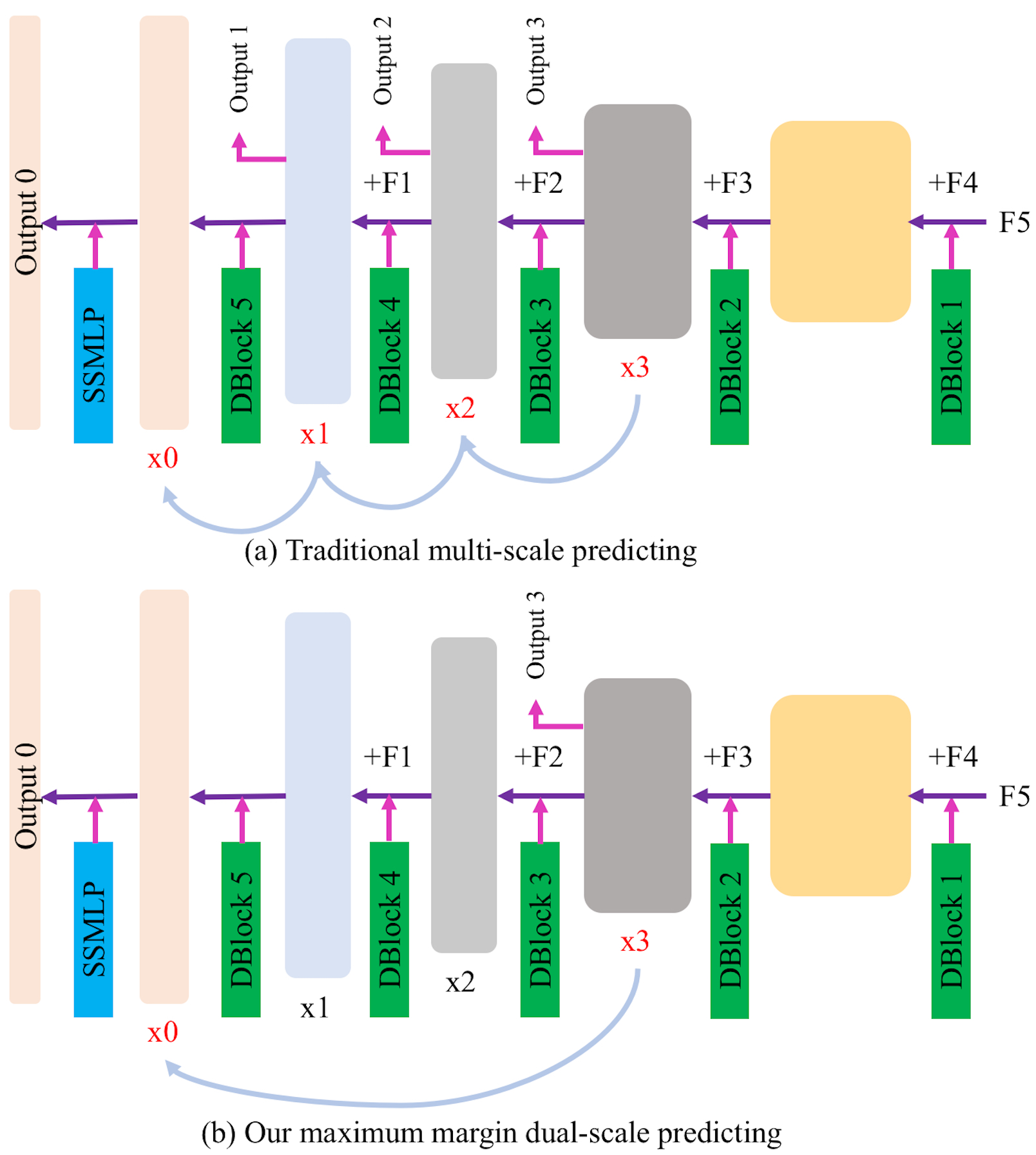}
    \caption{Illustrations of the conventional multi-scale prediction and the proposed maximum margin dual-scale prediction (MMDSP).}
    \label{dualscalepredicting}
\end{figure}

\begin{figure*}[htp]
    \centering
    \begin{overpic}[width=\textwidth]{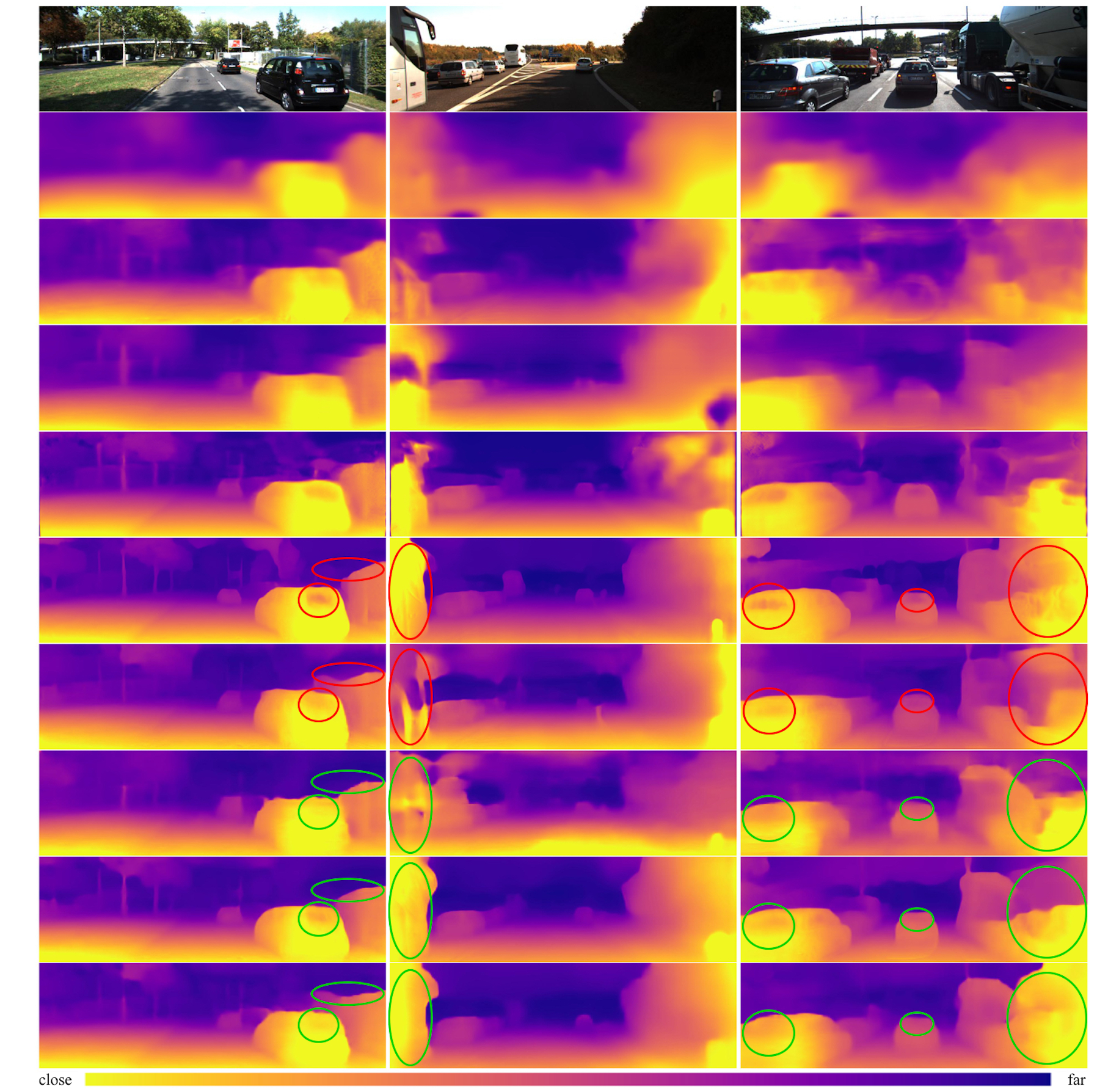}
    \put(0,92){\scriptsize\rotatebox{90}{Images}}
    \put(0,80){\scriptsize\rotatebox{90}{SfMLearner \cite{zhou2017unsupervised}}}
    \put(0,72){\scriptsize\rotatebox{90}{GeoNet \cite{yin2018geonet}}}
    \put(0,62){\scriptsize\rotatebox{90}{DDVO \cite{wang2018learning}}}    
    \put(0,51){\scriptsize\rotatebox{90}{Monodepth \cite{godard2017unsupervised}}}      
    \put(0,41){\scriptsize\rotatebox{90}{Monodepth2 \cite{godard2019digging}}} 
    \put(0,31){\scriptsize\rotatebox{90}{Monodepth2 \cite{godard2019digging}}}
    \put(1.5,31.2){\scriptsize\rotatebox{90}{w/o pretraining }}  
    \put(0,23){\scriptsize\rotatebox{90}{Our (LL)}}  
    \put(1.5,21.6){\scriptsize\rotatebox{90}{w/o pretraining}} 
    \put(0,14){\scriptsize\rotatebox{90}{Our (CC)}}  
    \put(1.5,12.7){\scriptsize\rotatebox{90}{w/ pretraining}} 
    \put(0,4){\scriptsize\rotatebox{90}{Our (CL)}}  
    \put(1.5,3.1){\scriptsize\rotatebox{90}{w/ pretraining}} 
    \end{overpic}
    \caption{Qualitative results on the KITTI dataset \cite{geiger2013vision}. Some reflective, far, and smooth regions, where prediction is difficult, are marked with red circles. Our model yields significantly improved predictions (green circles) for those regions. CC, CL, and LL represent the different network architectures, in which C and L represent convolutional neural networks (CNNs) and the proposed encoder/decoder. For instance, CL indicates that the encoder and decoder adopt CNNs and the proposed decoder, respectively.}
    \label{resultscomparison}
\end{figure*}
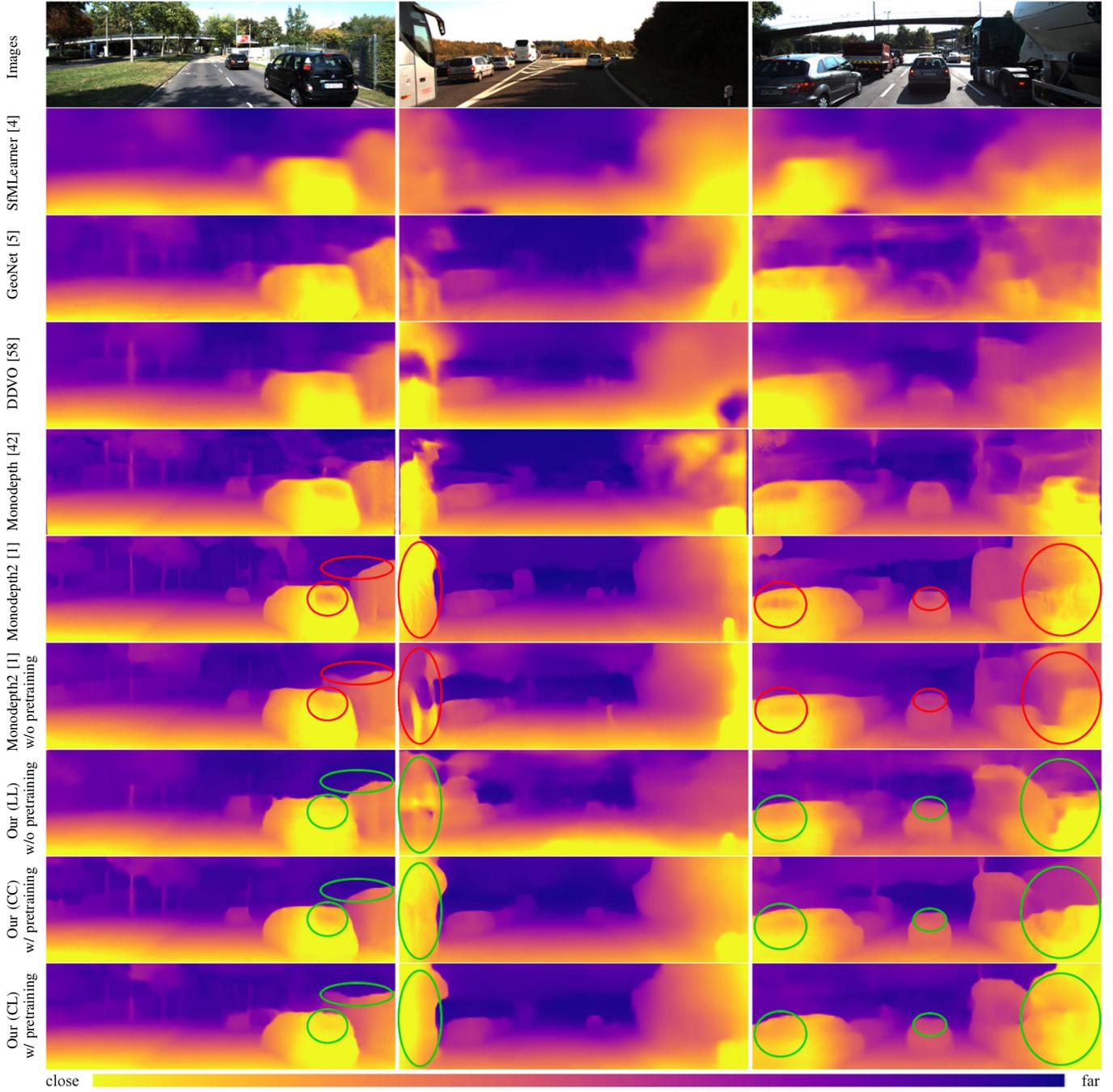

Intuitively, the network would extract the low-level vision features in the first several layers, whereas the deep features would contain more semantic information. For this case, consider the semantic level of the output and its previous feature to be $L_0$ and $L_1$, respectively. Then, the features $x0$, $x1$, $x2$, and $x3$, shown in Fig. \ref{dualscalepredicting} (a), would have the same semantic level of $L_1$. Therefore, four-scale prediction would require three times the feature transformations between features having the identical semantic level, with each transformation solely depending on a single decoder block; this approach increases the network’s learning difficulty.

Therefore, we propose the maximum margin dual-scale prediction (MMDSP) strategy to overcome the gradient locality issue, as shown in Fig. \ref{dualscalepredicting} (b). The proposed MMDSP only performs one transformation between features having the identical semantic level, with three decoder blocks.

The proposed MMDSP not only overcomes the gradient locality issue, thus improving performance, but also reduces the computational complexity. Ablation studies that experimentally demonstrate the effectiveness of the proposed approach are discussed in Section V-D.

\subsubsection{Final loss} We integrate the reconstruction loss, the proposed 3DGS loss, and the MMDSP to form the final loss (Eq. \ref{finalloss}) for training our networks:
\begin{equation}
\label{finalloss}
\mathcal{L}_{final} = \sum_{s=0,3} (\mathcal{L}_{recons} + \beta \frac{\mathcal{L}_{3DGS}}{2^{s}}),
\end{equation}
where $\beta$ is set to 0.001 following Godard's work \cite{godard2019digging}.

\begin{table*}[h]
\caption{Quantitative comparisons of depth estimation on the KITTI dataset \cite{geiger2013vision}. (B) indicates binocular/stereo input pairs, and (J) denotes joint learning of multiple tasks. GT, PT, and MS represent ground truth, pretraining, and model size, respectively. CC, CL, and LL represent the different network architectures, in which C and L represent convolutional neural networks (CNNs) and the proposed encoder/decoder. For instance, CL indicates that the encoder and decoder adopt CNNs and the proposed decoder, respectively. - represents that the situation is unclear. Notably, all self-supervised methods are trained at a resolution of $416\times128$ for a fair comparison. The best performances and our models are marked \textbf{bold}. The second best performances in the second and third cells are \underline{underlined}.}
\small
\centering
\begin{tabular}{@{}lcccccccccc@{}}
\toprule
{\color[HTML]{000000} } & {\color[HTML]{000000} } & {\color[HTML]{000000} } & {\color[HTML]{000000} } & \multicolumn{4}{c}{{\color[HTML]{000000} Errors $\downarrow$}} & \multicolumn{3}{c}{{\color[HTML]{000000} Errors $\uparrow$}} \\ \cmidrule(l){5-11} 
\multirow{-2}{*}{{\color[HTML]{000000} Methods}} & \multirow{-2}{*}{{\color[HTML]{000000} GT?}} & \multirow{-2}{*}{{\color[HTML]{000000} PT?}} & \multirow{-2}{*}{{\color[HTML]{000000} MS?}} & {\color[HTML]{000000} AbsRel} & {\color[HTML]{000000} SqRel} & {\color[HTML]{000000} RMS} & {\color[HTML]{000000} RMSlog} & {\color[HTML]{000000} $<1.25$} & {\color[HTML]{000000} $<1.25^{2}$} & {\color[HTML]{000000} $<1.25^{3}$} \\ \midrule
{\color[HTML]{000000} Eigen et al., coarse   \cite{eigen2014depth}} & \multicolumn{1}{c}{{\color[HTML]{000000} $\surd$}} & {\color[HTML]{000000} -} & {\color[HTML]{000000} -} & {\color[HTML]{000000} 0.214} & {\color[HTML]{000000} 1.605} & {\color[HTML]{000000} 6.563} & {\color[HTML]{000000} 0.292} & {\color[HTML]{000000} 0.673} & {\color[HTML]{000000} 0.884} & {\color[HTML]{000000} 0.957} \\
{\color[HTML]{000000} Eigen et al., fine \cite{eigen2014depth}} & \multicolumn{1}{c}{{\color[HTML]{000000} $\surd$}} & {\color[HTML]{000000} -} & {\color[HTML]{000000} -} & {\color[HTML]{000000} 0.203} & {\color[HTML]{000000} 1.548} & {\color[HTML]{000000} 6.307} & {\color[HTML]{000000} 0.282} & {\color[HTML]{000000} 0.702} & {\color[HTML]{000000} 0.890} & {\color[HTML]{000000} 0.958} \\
{\color[HTML]{000000} Liu et al. \cite{liu2015learning}} & \multicolumn{1}{c}{{\color[HTML]{000000} $\surd$}} & {\color[HTML]{000000} -} & {\color[HTML]{000000} -} & {\color[HTML]{000000} 0.202} & {\color[HTML]{000000} 1.614} & {\color[HTML]{000000} 6.523} & {\color[HTML]{000000} 0.275} & {\color[HTML]{000000} 0.678} & {\color[HTML]{000000} 0.895} & {\color[HTML]{000000} 0.965} \\
{\color[HTML]{000000} Kuznietsov et al. (B) \cite{kuznietsov2017semi}} & \multicolumn{1}{c}{{\color[HTML]{000000} $\surd$}} & {\color[HTML]{000000} $\surd$} & {\color[HTML]{000000} -} & {\color[HTML]{000000} 0.113} & {\color[HTML]{000000} 0.741} & {\color[HTML]{000000} 4.621} & {\color[HTML]{000000} 0.189} & {\color[HTML]{000000} 0.862} & {\color[HTML]{000000} 0.960} & {\color[HTML]{000000} 0.986} \\
{\color[HTML]{000000} DORN \cite{fu2018deep}} & \multicolumn{1}{c}{{\color[HTML]{000000} $\surd$}} & {\color[HTML]{000000} $\surd$} & {\color[HTML]{000000} $>$51.0M} & {\color[HTML]{000000} \textbf{0.072}} & {\color[HTML]{000000} \textbf{0.307}} & {\color[HTML]{000000} \textbf{2.727}} & {\color[HTML]{000000} \textbf{0.120}} & {\color[HTML]{000000} \textbf{0.932}} & {\color[HTML]{000000} \textbf{0.984}} & {\color[HTML]{000000} \textbf{0.994}} \\ \midrule
{\color[HTML]{000000} GeoNet-VGG (J) \cite{yin2018geonet}} & {\color[HTML]{000000} } & {\color[HTML]{000000} -} & {\color[HTML]{000000} -} & {\color[HTML]{000000} 0.164} & {\color[HTML]{000000} 1.303} & {\color[HTML]{000000} 6.090} & {\color[HTML]{000000} 0.247} & {\color[HTML]{000000} 0.765} & {\color[HTML]{000000} 0.919} & {\color[HTML]{000000} 0.968} \\
{\color[HTML]{000000} GeoNet-Resnet (J) \cite{yin2018geonet}} & {\color[HTML]{000000} } & {\color[HTML]{000000} -} & {\color[HTML]{000000} 229.3M} & {\color[HTML]{000000} 0.155} & {\color[HTML]{000000} 1.296} & {\color[HTML]{000000} 5.857} & {\color[HTML]{000000} 0.233} & {\color[HTML]{000000} 0.793} & {\color[HTML]{000000} 0.931} & {\color[HTML]{000000} 0.973} \\
{\color[HTML]{000000} DDVO \cite{wang2018learning}} & {\color[HTML]{000000} } & {\color[HTML]{000000} $\surd$} & {\color[HTML]{000000} -} & {\color[HTML]{000000} 0.151} & {\color[HTML]{000000} 1.257} & {\color[HTML]{000000} 5.583} & {\color[HTML]{000000} 0.228} & {\color[HTML]{000000} 0.810} & {\color[HTML]{000000} 0.936} & {\color[HTML]{000000} 0.974} \\
{\color[HTML]{000000} SC-SfMLearner \cite{bian2019unsupervised}} & {\color[HTML]{000000} } & {\color[HTML]{000000} $\surd$} & {\color[HTML]{000000} 59.4M} & {\color[HTML]{000000} 0.149} & {\color[HTML]{000000} 1.137} & {\color[HTML]{000000} 5.771} & {\color[HTML]{000000} 0.230} & {\color[HTML]{000000} 0.799} & {\color[HTML]{000000} 0.932} & {\color[HTML]{000000} 0.973} \\
{\color[HTML]{000000} Struct2depth \cite{casser2019depth}} & {\color[HTML]{000000} } & {\color[HTML]{000000} -} &{\color[HTML]{000000} -} &  {\color[HTML]{000000} \underline{0.141}} & {\color[HTML]{000000} 1.026} & {\color[HTML]{000000} 5.291} & {\color[HTML]{000000} 0.215} & {\color[HTML]{000000} 0.816} & {\color[HTML]{000000} 0.945} & {\color[HTML]{000000} 0.979} \\
{\color[HTML]{000000} Jia et al. \cite{mypaper2021}}& {\color[HTML]{000000} } & {\color[HTML]{000000} $\surd$} &{\color[HTML]{000000} 57.6M} & {\color[HTML]{000000} 0.144} & {\color[HTML]{000000} \textbf{0.966}} & {\color[HTML]{000000} 5.078} & {\color[HTML]{000000} 0.208} & {\color[HTML]{000000} 0.815} & {\color[HTML]{000000} 0.945} & {\color[HTML]{000000} \textbf{0.981}} \\
{\color[HTML]{000000} Monodepth2 \cite{godard2019digging}} & {\color[HTML]{000000} } & {\color[HTML]{000000} $\surd$} &{\color[HTML]{000000} 59.4M} &  {\color[HTML]{000000} \textbf{0.128}} & {\color[HTML]{000000} 1.087} & {\color[HTML]{000000} 5.171} & {\color[HTML]{000000} \underline{0.204}} & {\color[HTML]{000000} \textbf{0.855}} & {\color[HTML]{000000} 0.953} & {\color[HTML]{000000} 0.978} \\
{\color[HTML]{000000} \textbf{Our (CC)}}& {\color[HTML]{000000} } &{\color[HTML]{000000} $\surd$} & {\color[HTML]{000000} 59.4M} & {\color[HTML]{000000} \textbf{0.128}} & {\color[HTML]{000000} 0.990} & {\color[HTML]{000000} \underline{5.064}} & {\color[HTML]{000000} \textbf{0.202}} & {\color[HTML]{000000} \underline{0.851}} & {\color[HTML]{000000} \textbf{0.955}} & {\color[HTML]{000000} \underline{0.980}} \\
{\color[HTML]{000000} \textbf{Our (CL)}}& {\color[HTML]{000000} } &{\color[HTML]{000000} $\surd$} & {\color[HTML]{000000} 63.1M} & {\color[HTML]{000000} \textbf{0.128}} & {\color[HTML]{000000} \underline{0.979}} & {\color[HTML]{000000} \textbf{5.033}} & {\color[HTML]{000000} \textbf{0.202}} & {\color[HTML]{000000} \underline{0.851}} & {\color[HTML]{000000} \underline{0.954}} & {\color[HTML]{000000} \underline{0.980}} \\ \midrule
{\color[HTML]{000000} SfMLearner \cite{zhou2017unsupervised}} & {\color[HTML]{000000} } & {\color[HTML]{000000} } & {\color[HTML]{000000} 126.0M} & {\color[HTML]{000000} 0.208} & {\color[HTML]{000000} 1.768} & {\color[HTML]{000000} 6.856} & {\color[HTML]{000000} 0.283} & {\color[HTML]{000000} 0.678} & {\color[HTML]{000000} 0.885} & {\color[HTML]{000000} 0.957} \\
{\color[HTML]{000000} Yang et al. (J) \cite{yang2017unsupervised}} & {\color[HTML]{000000} } & {\color[HTML]{000000} } &{\color[HTML]{000000} 126.0M} &  {\color[HTML]{000000} 0.182} & {\color[HTML]{000000} 1.481} & {\color[HTML]{000000} 6.501} & {\color[HTML]{000000} 0.267} & {\color[HTML]{000000} 0.725} & {\color[HTML]{000000} 0.906} & {\color[HTML]{000000} 0.963} \\
{\color[HTML]{000000} Monodepth2 \cite{godard2019digging}} & {\color[HTML]{000000} } & {\color[HTML]{000000} } &{\color[HTML]{000000} 59.4M} & {\color[HTML]{000000} \underline{0.144}} & {\color[HTML]{000000} \textbf{1.059}} & {\color[HTML]{000000} \underline{5.289}} & {\color[HTML]{000000} \underline{0.217}} & {\color[HTML]{000000} \underline{0.824}} & {\color[HTML]{000000} \textbf{0.945}} & {\color[HTML]{000000} \underline{0.976}} \\
{\color[HTML]{000000} \textbf{Our (LL)}}& {\color[HTML]{000000} } &{\color[HTML]{000000} } & {\color[HTML]{000000} \textbf{25.8M}} & {\color[HTML]{000000} \textbf{0.141}} & {\color[HTML]{000000} \underline{1.060}} & {\color[HTML]{000000} \textbf{5.247}} & {\color[HTML]{000000} \textbf{0.215}} & {\color[HTML]{000000} \textbf{0.830}} & {\color[HTML]{000000} \underline{0.944}} & {\color[HTML]{000000} \textbf{0.977}} \\
\bottomrule
\end{tabular}
\label{kittiresults}
\end{table*}

\section{Experiments}
In this section, we first introduce the experiment implementation details and then compare the results of the proposed approach with those of other state-of-the-art methods. Thereafter, the parameter analysis, ablation studies, and model complexity are discussed. All experiments are implemented with PyTorch $1.5.0$ library on a single GTX $1080$Ti GPU card.

\begin{table*}[htb]
\caption{Quantitative comparisons of depth estimation on the Make3D dataset \cite{saxena2008make3d}. CC, CL, and LL represent the different network architectures, in which C and L represent convolutional neural networks (CNNs) and the proposed encoder/decoder. For instance, CL indicates that the encoder and decoder adopt CNNs and the proposed decoder, respectively. The best performances and our models are marked \textbf{bold}. The second best performances in the second box are \underline{underlined}.}
\small
\centering
\begin{tabular}{@{}lcccccc@{}}
\toprule
{\color[HTML]{000000} } & {\color[HTML]{000000} } & {\color[HTML]{000000} } & \multicolumn{4}{c}{{\color[HTML]{000000} Errors $\downarrow$}} \\ \cmidrule(l){4-7} 
\multirow{-2}{*}{{\color[HTML]{000000} Methods}} & \multirow{-2}{*}{{\color[HTML]{000000} GT?}} & \multirow{-2}{*}{{\color[HTML]{000000} PT?}} & {\color[HTML]{000000} AbsRel} & {\color[HTML]{000000} SqRel} & {\color[HTML]{000000} RMS} & {\color[HTML]{000000} RMSlog} \\ \midrule
{\color[HTML]{000000} Karsch et al. \cite{karsch2014depth}} & \multicolumn{1}{c}{{\color[HTML]{000000} $\surd$}} & {\color[HTML]{000000} } & {\color[HTML]{000000} 0.428} & {\color[HTML]{000000} 5.079} & {\color[HTML]{000000} 8.389} & {\color[HTML]{000000} 0.149} \\
{\color[HTML]{000000} Liu et al. \cite{liu2014discrete}} & \multicolumn{1}{c}{{\color[HTML]{000000} $\surd$}} & {\color[HTML]{000000} } & {\color[HTML]{000000} 0.475} & {\color[HTML]{000000} 6.562} & {\color[HTML]{000000} 10.05} & {\color[HTML]{000000} 0.165} \\
{\color[HTML]{000000} Laina et al. \cite{laina2016deeper}} & \multicolumn{1}{c}{{\color[HTML]{000000} $\surd$}} & {\color[HTML]{000000} $\surd$} & {\color[HTML]{000000} \textbf{0.204}} & {\color[HTML]{000000} \textbf{1.840}} & {\color[HTML]{000000} \textbf{5.683}} & {\color[HTML]{000000} \textbf{0.084}} \\
\midrule
{\color[HTML]{000000} DDVO \cite{wang2018learning}} & {\color[HTML]{000000} } & {\color[HTML]{000000} $\surd$} & {\color[HTML]{000000} 0.387} & {\color[HTML]{000000} 4.720} & {\color[HTML]{000000} 8.090} & {\color[HTML]{000000} 0.204} \\
{\color[HTML]{000000} Monodepth \cite{godard2017unsupervised}} & {\color[HTML]{000000} } & {\color[HTML]{000000} $\surd$} & {\color[HTML]{000000} 0.544} & {\color[HTML]{000000} 10.94} & {\color[HTML]{000000} 11.760} & {\color[HTML]{000000} \underline{0.193}} \\
{\color[HTML]{000000} Monodepth2 \cite{godard2019digging}} & {\color[HTML]{000000} } & {\color[HTML]{000000} $\surd$} & {\color[HTML]{000000} 0.322} & {\color[HTML]{000000} 3.589} & {\color[HTML]{000000} 7.417} & {\color[HTML]{000000} \textbf{0.163}} \\
{\color[HTML]{000000} Jia et al. \cite{mypaper2021}}& {\color[HTML]{000000} } & {\color[HTML]{000000} $\surd$} & {\color[HTML]{000000} 0.301} & {\color[HTML]{000000} 3.143} & {\color[HTML]{000000} 6.972} & {\color[HTML]{000000} 0.351} \\
{\color[HTML]{000000}  \textbf{Our (CL)}}& {\color[HTML]{000000} } & {\color[HTML]{000000} $\surd$ } & {\color[HTML]{000000} \underline{0.269}} & {\color[HTML]{000000} \underline{2.201}} & {\color[HTML]{000000} \underline{6.452}} & {\color[HTML]{000000} 0.325} \\
{\color[HTML]{000000}  \textbf{Our (CC)}}& {\color[HTML]{000000} } & {\color[HTML]{000000} $\surd$ } & {\color[HTML]{000000} \textbf{0.267}} & {\color[HTML]{000000} \textbf{2.188}} & {\color[HTML]{000000} \textbf{6.406}} & {\color[HTML]{000000} 0.322} \\
\midrule
{\color[HTML]{000000} SfMLearner \cite{zhou2017unsupervised}} & {\color[HTML]{000000} } & {\color[HTML]{000000} } & {\color[HTML]{000000} 0.383} & {\color[HTML]{000000} 5.321} & {\color[HTML]{000000} 10.470} & {\color[HTML]{000000} 0.478} \\
{\color[HTML]{000000}  \textbf{Our (LL)}}& {\color[HTML]{000000} } & {\color[HTML]{000000} } & {\color[HTML]{000000} \textbf{0.289}} & {\color[HTML]{000000} \textbf{2.423}} & {\color[HTML]{000000} \textbf{6.701}} & {\color[HTML]{000000} \textbf{0.348}} \\
\bottomrule
\end{tabular}
\label{make3dresults}
\end{table*}

\begin{table*}[h]
\centering
\caption{Experiments on different $k$ values. The best performances are marked \textbf{bold}.}
\begin{tabular}{lccccccc}
\toprule
\multirow{2}{*}{$k$} & \multicolumn{4}{c}{Errors $\downarrow$} & \multicolumn{3}{c}{Errors $\uparrow$} \\ \cline{2-8} 
 & AbsRel & SqRel & RMS & RMSlog &  $<1.25$ & $<1.25^{2}$ & $<1.25^{3}$ \\ \hline
32 & \textbf{0.142} & 1.121 & 5.330 & \textbf{0.216} & \textbf{0.829} & \textbf{0.944} & 0.976 \\
64 & \textbf{0.142} & \textbf{1.094} & \textbf{5.286} & \textbf{0.216} & 0.827 & \textbf{0.944} & \textbf{0.977} \\
128 & 0.143 & 1.117 & 5.340 & 0.217 & \textbf{0.829} & 0.942 & 0.975 \\ \bottomrule
\end{tabular}
\label{kvalues}
\end{table*}

\begin{table*}[h]
\centering
\caption{Experiments on different $one\_kv\_heads$. The best performances are marked \textbf{bold}.}
\begin{tabular}{lccccccc}
\toprule
\multirow{2}{*}{$one\_kv\_heads$} & \multicolumn{4}{c}{Errors $\downarrow$} & \multicolumn{3}{c}{Errors $\uparrow$} \\ \cline{2-8} 
 & AbsRel & SqRel & RMS & RMSlog & $<1.25$ & $<1.25^{2}$ & $<1.25^{3}$ \\ \hline
False & 0.144 & 1.106 & 5.339 & 0.217 & 0.825 & 0.943 & 0.976 \\
True & \textbf{0.142} & \textbf{1.094} & \textbf{5.286} & \textbf{0.216} & \textbf{0.827} & \textbf{0.944} & \textbf{0.977} \\
 \bottomrule
\end{tabular}
\label{onekvheads}
\end{table*}

\subsection{Implementation details}

\textbf{Datasets.} KITTI \cite{geiger2013vision}, a large-scale and publicly available dataset widely used in various computer vision tasks, serves as the set of benchmarks for evaluation and comparison in this study. Following Zhou et al. \cite{zhou2017unsupervised}, we take  $40,109$ and $4,431$ $3$-frame sequences from the KITTI dataset as the training and validation data, respectively. Following Eigen et al.’s \cite{eigen2014depth} split, we take $697$ images from the KITTI dataset for testing.

We apply our model trained on the KITTI dataset to $134$ test images of the Make3D dataset \cite{saxena2008make3d} unseen during training for demonstrating the generalization capability of the proposed model. Following Godard et al. \cite{godard2019digging}, we evaluate Make3D’s images on a center crop at a $2\times1$ ratio. Additionally, qualitative results on the Cityscapes dataset \cite{cordts2016cityscapes} generated by the proposed model trained on the KITTI dataset are presented. In all experiments, the resolution of the input image is $416 \times 128$.


\textbf{Data augmentation.} Following Godard et al. \cite{godard2019digging}, the input images are augmented with random cropping, scaling, and horizontal flips. In addition, a set of color augmentations, random brightness, contrast, saturation, and hue jitter with respective ranges of $\pm 0.2$, $\pm 0.2$, $\pm 0.2$, and $\pm 0.1$ are adopted with a $50$ percent chance. These color augmentations are solely applied to the images fed to the networks, not to those applied to compute loss.

\textbf{Hyperparameters.} Our networks are trained using the Adam optimizer \cite{kingma2014adam}. The learning rate is initially set to $10^{-4}$ and decreased by a factor of $10$ every $15$ epochs. The other parameters of the optimizer are set to the default values. The epoch, batch size, and length of the sequence are set to $25$, $12$, and $3$, respectively. 

During training, we initialize the weights and biases of a linear layer using a Gaussian distribution with standard deviation $0.02$ and constant $0$. For the LayerNorm layer, the weights and biases are initialized with constants $1$ and $0$, respectively. We do not pretrain our model on the ImageNet dataset because of computational resource constraints. The specific network configurations are presented in Section V-C.

\subsection{Results}
Table \ref{kittiresults} shows the quantitative results of the proposed model on the KITTI dataset \cite{geiger2013vision}, divided into three categories according to whether the ground truth and pretraining are used during the training phase. Of note is that the depth estimation results in this paper benefit from the fusion scaling strategy proposed in Jia's work \cite{mypaper2021}.

To benefit from pretraining and for a fair comparison, we integrate the proposed loss and decoders with existing pre-trained CNNs. The first network architecture CC, in which the networks are identical to those in Monodepth2 \cite{godard2019digging} but apply the proposed loss function, outperforms the original Monodepth2, demonstrating the effectiveness of the proposed loss function. Then, we replace the CNN decoders with the proposed decoders, forming the second network architecture CL, which further improves the performance. Finally, the LL network, consisting of the proposed DLNet (encoder) and decoders, achieves performance competitive to those of the state-of-the-art methods while also reducing the number of parameters by more than $56\%$.

Figure \ref{performance} presents a comprehensive comparison of the performance and number of parameters. Without pretraining, the proposed model (LL) outperforms the other methods, with performance comparable even to those of pretrained models. Moreover, the proposed model significantly reduces the number of parameters, and simply integrating the proposed loss function and decoders with the pre-trained encoder used in previous works leads to better performance. 

Figure \ref{resultscomparison} illustrates the promising qualitative results of the proposed model on the KITTI dataset. The regions marked with red circles are extremely challenging to accurately predict because of the issues of reflection, smoothness, and farness. However, effectively capturing global features using CNNs is difficult and may lead to unexpected failure cases, especially for a model without pretraining. Thereagainst, the proposed loss and networks implicitly impose geometry constraints on the results and effectively extract the global information, respectively, significantly improving the performance and contributing to the prediction of a more accurate depth map.

\begin{table*}[h]
\centering
\caption{Experiments on different $share\_kv$. The best performances are marked \textbf{bold}.}
\begin{tabular}{lccccccc}
\toprule
\multirow{2}{*}{$share\_kv$} & \multicolumn{4}{c}{Errors $\downarrow$} & \multicolumn{3}{c}{Errors $\uparrow$} \\ \cline{2-8} 
 & AbsRel & SqRel & RMS & RMSlog & $<1.25$ & $<1.25^{2}$ & $<1.25^{3}$ \\ \hline
Flase & 0.144 & 1.122 & \textbf{5.285} & 0.217 & \textbf{0.827} & \textbf{0.944} & 0.976 \\
True & \textbf{0.142} & \textbf{1.094} & 5.286 & \textbf{0.216} & \textbf{0.827} & \textbf{0.944} & \textbf{0.977} \\ \bottomrule
\end{tabular}
\label{sharekv}
\end{table*}

\begin{table*}[h]
\centering
\caption{Experiments on different $heads$. The best performances are marked \textbf{bold}.}
\begin{tabular}{lccccccc}
\toprule
\multirow{2}{*}{$heads$} & \multicolumn{4}{c}{Errors $\downarrow$} & \multicolumn{3}{c}{Errors $\uparrow$} \\ \cline{2-8} 
 & AbsRel & SqRel & RMS & RMSlog & $<1.25$ & $<1.25^{2}$ & $<1.25^{3}$ \\ \hline
4 & 0.144 & 1.130 & 5.317 & 0.217 & 0.827 & 0.943 & 0.976 \\
8 & \textbf{0.142} & \textbf{1.094} & 5.286 & \textbf{0.216} & 0.827 & \textbf{0.944} & \textbf{0.977} \\
16 & 0.143 & 1.114 & \textbf{5.282} & \textbf{0.216} & \textbf{0.828} & \textbf{0.944} & \textbf{0.977} \\ \bottomrule
\end{tabular}
\label{heads}
\end{table*}

\begin{table*}[h]
\centering
\caption{Experiments on different activation functions. The best performances are marked \textbf{bold}.}
\begin{tabular}{lccccccc}
\toprule
\multirow{2}{*}{activation functions} & \multicolumn{4}{c}{Errors $\downarrow$} & \multicolumn{3}{c}{Errors $\uparrow$} \\ \cline{2-8} 
 & AbsRel & SqRel & RMS & RMSlog & $<1.25$ & $<1.25^{2}$ & $<1.25^{3}$ \\ \hline
ReLu \cite{relu} & 0.145 & 1.103 & 5.291 & 0.218 & 0.822 & 0.943 & 0.976 \\
ELU \cite{elu} & 0.146 & 1.141 & 5.353 & 0.219 & 0.823 & 0.942 & 0.975 \\
GELU \cite{gelu} & \textbf{0.142} & \textbf{1.094} & \textbf{5.286} & \textbf{0.216} & \textbf{0.827} & \textbf{0.944} & \textbf{0.977} \\ \bottomrule
\end{tabular}
\label{activationfunctions}
\end{table*}

\begin{table*}[h]
\centering
\caption{Experiments on different $c$. MS represents model size. The best performances are marked \textbf{bold}.}
\begin{tabular}{lcccccccc}
\toprule
\multirow{2}{*}{$c$} & \multirow{2}{*}{MS} & \multicolumn{4}{c}{Errors $\downarrow$} & \multicolumn{3}{c}{Errors $\uparrow$} \\ \cline{3-9} 
 & & AbsRel & SqRel & RMS & RMSlog & $<1.25$ & $<1.25^{2}$ & $<1.25^{3}$ \\ \hline
0.5 & 58.0M & \textbf{0.140} & 1.079 & 5.348 & 0.215 & \textbf{0.831} & \textbf{0.944} & 0.976 \\
1  & 25.8M & 0.141 & \textbf{1.060} & \textbf{5.247} & \textbf{0.215} & 0.830 & \textbf{0.944} & \textbf{0.977} \\
2  & 15.1M & 0.146 & 1.135 & 5.374 & \textbf{0.218} & 0.821 & 0.941 & 0.976 \\ \bottomrule
\end{tabular}
\label{hiddendimensionresults}
\end{table*}

\begin{table*}[h]
\centering
\caption{Experiments on different Linformer \cite{Linformer} depths. MS represents model size. The best performances are marked \textbf{bold}.}
\begin{tabular}{lcccccccc}
\toprule
\multirow{2}{*}{Linformer depth} & \multirow{2}{*}{MS} & \multicolumn{4}{c}{Errors $\downarrow$} & \multicolumn{3}{c}{Errors $\uparrow$} \\ \cline{3-9} 
 & & AbsRel & SqRel & RMS & RMSlog & $<1.25$ & $<1.25^{2}$ & $<1.25^{3}$ \\ \hline
1  & 25.8M & \textbf{0.141} & \textbf{1.060} & \textbf{5.247} & \textbf{0.215} & \textbf{0.830} & \textbf{0.944} & \textbf{0.977} \\
2  & 30.7M & 0.144 & 1.111 & 5.299 & 0.217 & 0.826 & 0.943 & 0.976 \\ \bottomrule
\end{tabular}
\label{differentLinformerlayers}
\end{table*}

\begin{table*}[h]
\centering
\caption{Experiments on different number of DLBlock. MS represents model size. The best performances are marked \textbf{bold}.}
\begin{tabular}{lcccccccc}
\toprule
\multirow{2}{*}{Number of DLBlock} & \multirow{2}{*}{MS} & \multicolumn{4}{c}{Errors $\downarrow$} & \multicolumn{3}{c}{Errors $\uparrow$} \\ \cline{3-9} 
 & & AbsRel & SqRel & RMS & RMSlog & $<1.25$ & $<1.25^{2}$ & $<1.25^{3}$ \\ \hline
1  & 25.8M & \textbf{0.141} & \textbf{1.060} & \textbf{5.247} & \textbf{0.215} & \textbf{0.830} & \textbf{0.944} & \textbf{0.977} \\
2  & 34.7M & 0.144 & 1.171 & 5.323 & 0.218 & 0.831 & 0.944 & 0.975 \\ \bottomrule
\end{tabular}
\label{differentnumberofDLBlock}
\end{table*}

\begin{table*}[htp]
\centering
\caption{Experiments on different multi-scale prediction strategies. The best performances are marked \textbf{bold}. The scale numbers are consistent with the numbers in Fig. \ref{dualscalepredicting}, in which 0 and 3 are our maximum margin dual-scale prediction (MMDSP).}
\begin{tabular}{lccccccc}
\toprule
\multirow{2}{*}{scales} & \multicolumn{4}{c}{Errors $\downarrow$} & \multicolumn{3}{c}{Errors $\uparrow$} \\ \cline{2-8} 
 & AbsRel & SqRel & RMS & RMSlog & $<1.25$ & $<1.25^{2}$ & $<1.25^{3}$ \\ \hline
0 & 0.143 & \textbf{1.103} & 5.347 & 0.217 & 0.828 & \textbf{0.944} & \textbf{0.976} \\
0,3 (MMDSP) & \textbf{0.142} & 1.118 & \textbf{5.330} & \textbf{0.216} & \textbf{0.829} & \textbf{0.944} & \textbf{0.976} \\
0,2,3 & 0.145 & 1.184 & 5.448 & 0.220 & 0.822 & 0.942 & \textbf{0.976} \\
0,1,2,3 & 0.144 & 1.115 & 5.347 & 0.217 & 0.823 & 0.943 & \textbf{0.976} \\ \bottomrule
\end{tabular}
\label{multiscalepredicting}
\end{table*}

\begin{table*}[htp]
\caption{Ablation studies on loss function. B, MRp, SSIM, AM, MMDSP, and 3DGS denote the basic photometric loss, minimal reprojection, SSIM loss, automasking, proposed maximum margin dual-scale prediction, and 3D geometry smoothness loss, respectively. The best performances are marked \textbf{bold}.}
\small
\centering
\begin{tabular}{@{}lccccccc@{}}
\toprule
{\color[HTML]{000000} } & \multicolumn{4}{c}{{\color[HTML]{000000} Errors $\downarrow$}} & \multicolumn{3}{c}{{\color[HTML]{000000} Errors $\uparrow$}} \\ \cmidrule(l){2-8} 
\multirow{-2}{*}{{\color[HTML]{000000} Methods}} & {\color[HTML]{000000} AbsRel} & {\color[HTML]{000000} SqRel} & {\color[HTML]{000000} RMS} & {\color[HTML]{000000} RMSlog} & {\color[HTML]{000000} $<1.25$} & {\color[HTML]{000000} $<1.25^{2}$} & {\color[HTML]{000000} $<1.25^{3}$} \\ \midrule
{\color[HTML]{000000} B} & {\color[HTML]{000000} 0.218} & {\color[HTML]{000000} 4.062} & {\color[HTML]{000000} 6.788} & {\color[HTML]{000000} 0.286} & {\color[HTML]{000000} 0.772} & {\color[HTML]{000000} 0.915} & {\color[HTML]{000000} 0.959} \\
{\color[HTML]{000000} B+MRp} & {\color[HTML]{000000} 0.194} & {\color[HTML]{000000} 2.756} & {\color[HTML]{000000} 6.213} & {\color[HTML]{000000} 0.262} & {\color[HTML]{000000} 0.786} & {\color[HTML]{000000} 0.925} & {\color[HTML]{000000} 0.966} \\
{\color[HTML]{000000} B+MRp+SSIM} & {\color[HTML]{000000} 0.148} & {\color[HTML]{000000} 1.238} & {\color[HTML]{000000} 5.496} & {\color[HTML]{000000} 0.221} & {\color[HTML]{000000} 0.818} & {\color[HTML]{000000} 0.941} & {\color[HTML]{000000} 0.974} \\
{\color[HTML]{000000} B+MRp+SSIM+AM} & {\color[HTML]{000000} 0.144} & {\color[HTML]{000000} 1.120} & {\color[HTML]{000000} 5.308} & {\color[HTML]{000000} 0.217} & {\color[HTML]{000000} 0.826} & {\color[HTML]{000000} \textbf{0.944}} & {\color[HTML]{000000} 0.976} \\
{\color[HTML]{000000} B+MRp+SSIM+AM+MMDSP} & {\color[HTML]{000000} 0.142} & {\color[HTML]{000000} 1.118} & {\color[HTML]{000000} 5.330} & {\color[HTML]{000000} 0.216} & {\color[HTML]{000000} 0.829} & {\color[HTML]{000000} \textbf{0.944}} & {\color[HTML]{000000} 0.976} \\
{\color[HTML]{000000} B+MRp+SSIM+AM+MMDSP+3DGS} & {\color[HTML]{000000} \textbf{0.141}} & {\color[HTML]{000000} \textbf{1.060}} & {\color[HTML]{000000} \textbf{5.247}} & {\color[HTML]{000000} \textbf{0.215}} & {\color[HTML]{000000} \textbf{0.830}} & {\color[HTML]{000000} \textbf{0.944}} & {\color[HTML]{000000} \textbf{0.977}} \\ \bottomrule
\end{tabular}
\label{lossfunctions}
\end{table*}

\begin{figure}[htp]
    \centering
    \begin{overpic}[scale=0.5]{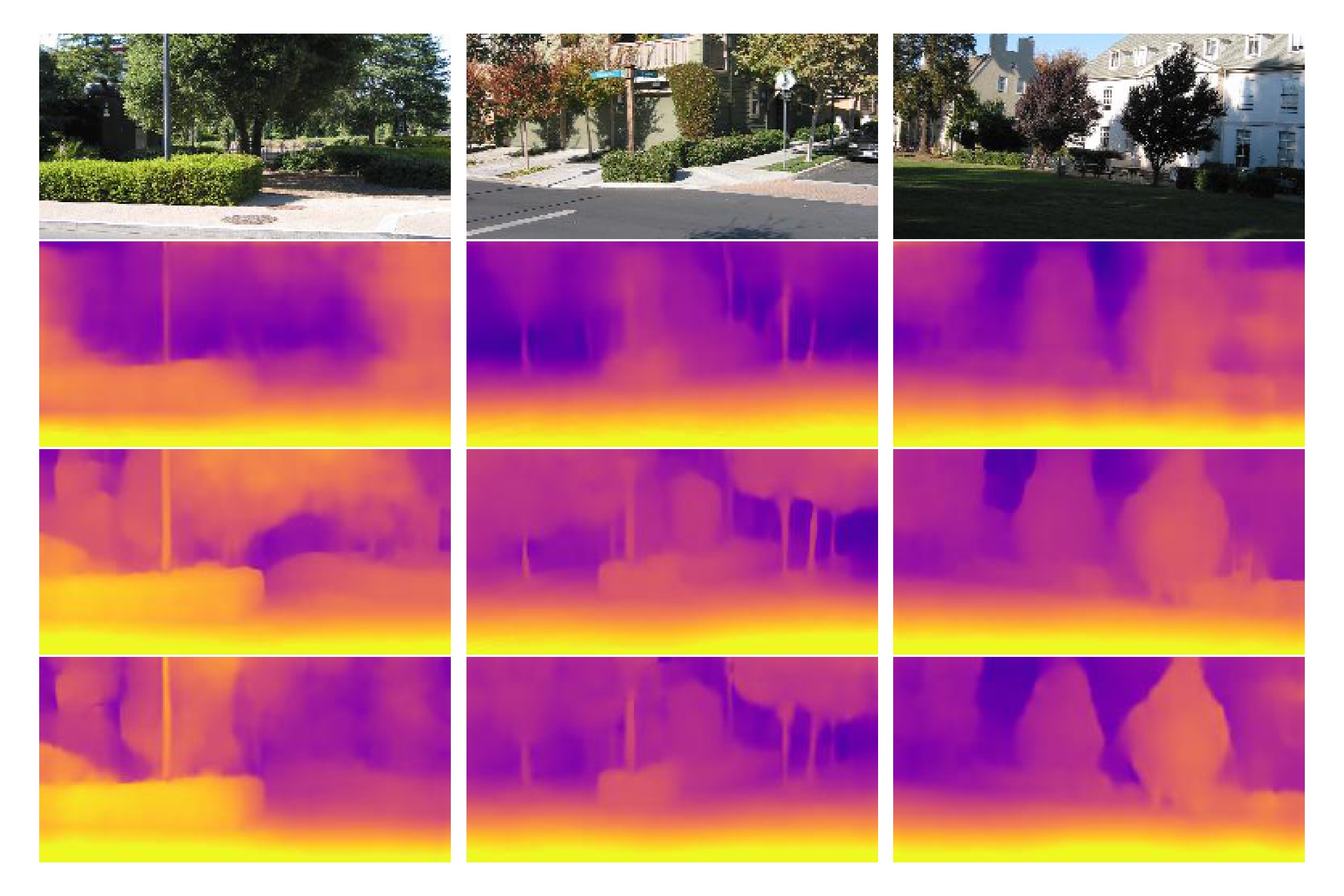}
    \put(-2.5,52){\tiny\rotatebox{90}{Images}}
    \put(-2.5,36){\tiny\rotatebox{90}{Our (LL)}}  
    \put(-0.5,33.1){\tiny\rotatebox{90}{w/o pretraining}} 
    \put(-2.5,21){\tiny\rotatebox{90}{Our (CC)}}  
    \put(-0.5,18.1){\tiny\rotatebox{90}{w/ pretraining}} 
    \put(-2.5,6){\tiny\rotatebox{90}{Our (CL)}}  
    \put(-0.5,3.1){\tiny\rotatebox{90}{w/ pretraining}} 
    \end{overpic}
    \caption{Qualitative results on the Make3D dataset \cite{saxena2008make3d}. CC, CL, and LL represent the different network architectures, in which C and L represent convolutional neural networks (CNNs) and the proposed encoder/decoder. For instance, CL indicates that the encoder and decoder adopt CNNs and the proposed decoder, respectively. Note that we directly apply the model trained on the KITTI dataset \cite{geiger2013vision} to the Make3D dataset \cite{saxena2008make3d}, without any refinements.}
    \label{make3dqualitative}
\end{figure}

To evaluate the generalization capability of the proposed model, we directly apply our model trained on the KITTI dataset to the Make3D dataset without any refinements and training. Table \ref{make3dresults} shows that the quantitative results output by the proposed model are competitive, regardless of pretraining status.

Figure \ref{make3dqualitative} illustrates some qualitative results generated by the proposed model on the Make3D dataset, which indicate that the proposed model has powerful generalization capability and can predict a reasonable depth map with adequate details for unseen data. The scenarios and perspectives in Make3D differ substantially from those in the KITTI data used for training.

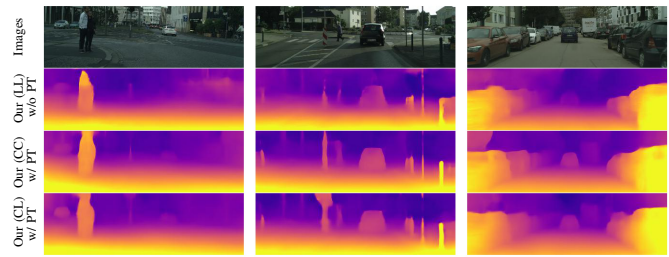
\begin{figure}[htp]
    \centering
    \begin{overpic}[scale=0.35]{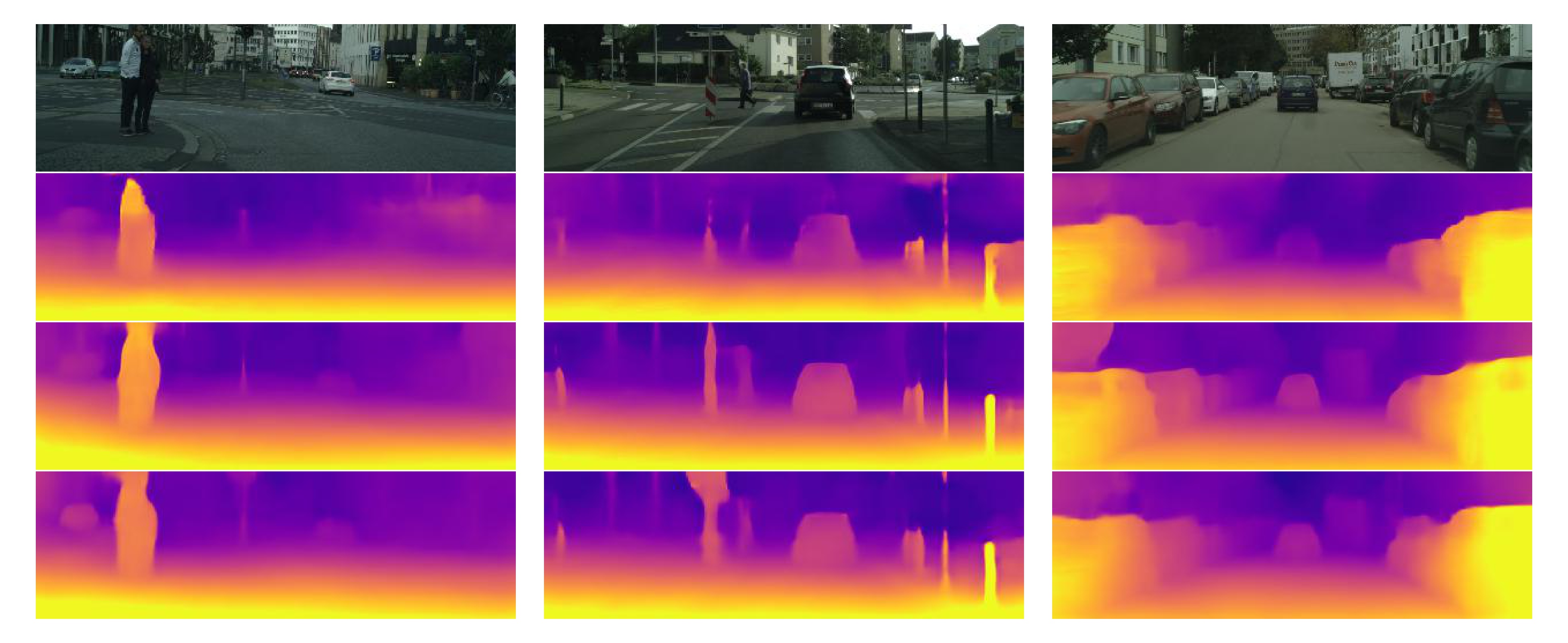}
    \put(-2.5,32){\tiny\rotatebox{90}{Images}}
    \put(-2.5,21.6){\tiny\rotatebox{90}{Our (LL)}}  
    \put(-0.5,22.6){\tiny\rotatebox{90}{w/o PT}} 
    \put(-2.5,11.6){\tiny\rotatebox{90}{Our (CC)}}  
    \put(-0.5,12.6){\tiny\rotatebox{90}{w/ PT}} 
    \put(-2.5,2.6){\tiny\rotatebox{90}{Our (CL)}}  
    \put(-0.5,3.6){\tiny\rotatebox{90}{w/ PT}} 
    \end{overpic}
    \caption{Qualitative results on the Cityscapes dataset \cite{cordts2016cityscapes}. CC, CL, and LL represent the different network architectures, in which C and L represent convolutional neural networks (CNNs) and the proposed encoder/decoder. For instance, CL indicates that the encoder and decoder adopt CNNs and the proposed decoder, respectively. Note that we directly apply the model trained on the KITTI dataset \cite{geiger2013vision} to the Cityscapes dataset \cite{cordts2016cityscapes}, without any refinements. PT indicates pretraining.}
    \label{Cityscapesqualitative}
\end{figure}

In Fig. \ref{Cityscapesqualitative}, we present some qualitative results predicted by the proposed model on the Cityscapes dataset, thus demonstrating the model’s practicality. The scenes, color distributions, and perspectives in Cityscapes differ from those in the KITTI data used for training.

Summarily, we can conclude that the proposed model achieves competitive performance on the KITTI, Make3D, and Cityscapes datasets, presenting excellent qualitative results, especially in some challenging regions.

\subsection{Parameter analysis}
In this part, we report on exhaustive experiments on the different parameters of the proposed model, conducted to determine the optimal configurations.

Primarily, we conduct a series of experiments to search for the best set of parameters for the Linformer block embedded in our networks. As shown in Tables \ref{kvalues}, \ref{onekvheads}, \ref{sharekv}, and \ref{heads}, we carefully perform the experiments for each parameter in the Linformer \cite{Linformer} block and choose the parameters that result in a better performance as the final configurations. Specifically, we set the parameter $k$, $one\_kv\_heads$, $share\_kv$, and $heads$ to $64$, $True$, $True$, and $8$, respectively.

We further experiment with various global network parameters. As shown in Table \ref{activationfunctions}, we evaluate the different activation functions and find that GELU \cite{gelu} outperforms other activation functions by a clear margin. Table \ref{hiddendimensionresults} summarizes the performance of different hidden dimensions defined in Eq. \ref{hiddendimension}. Considering both performance and model efficiency, $c$ is set to $1$.

Finally, the results for experiments on networks with various Linformer depths and DLBlock amounts are shown in Table \ref{differentLinformerlayers} and \ref{differentnumberofDLBlock}. No performance improvements could be achieved by deepening the network. Therefore, the final networks adopt a single DLBlock in which the Linformer depth is set as $1$ for each encoder block. All results reported in Section V-B are derived from these configurations.

\subsection{Ablation studies}
In this section, we discuss experiments conducted to validate the proposed MMDSP strategy and loss items. First, Table \ref{multiscalepredicting} shows that the proposed MMDSP is more effective than the other multi-scale prediction strategies. In fact, the proposed Linformer-based networks are capable of concurrently capturing the global and local features, which contribute to overcoming the gradient locality issue. Therefore, the performance of our network with a single prediction scale is still on par with that of the network equipped with the proposed MMDSP.

Table \ref{lossfunctions} demonstrates the effectiveness of each item of the proposed loss function, which indicates that the proposed MMDSP strategy and especially the 3DGS loss indeed improve the performance.

\subsection{Model complexity}
Table \ref{modelinfo} summarizes the model’s time and space complexities. Clearly, the proposed model outperforms the other state-of-the-art methods by large margins in both time and space complexities.

For reference, our depth and pose networks can achieve a speed of more than $88$ and $172$ frames per second at a resolution of $416 \times 128$ on a single GTX $1080$Ti card (averaging over $100$ iterations), respectively, which can satisfy the requirements of most applications.

\begin{table}[htp]
\centering
\caption{Model complexity. The time complexity evaluations are performed at an image resolution of $416 \times 128$ on a single GTX $1080$Ti card. TC and SC represent time and space complexities, respectively. The best performances are marked \textbf{bold}.}
\begin{tabular}{lcc}
\toprule
Methods & TC (GFLOPs) & SC (M) \\  \midrule
Jia et al. \cite{mypaper2021} & 3.485 & 57.6 \\
Monodepth2 \cite{godard2019digging} &  3.480 & 59.4 \\ 
Our & \textbf{1.311} & \textbf{25.8} \\
\bottomrule
\end{tabular}
\label{modelinfo}
\end{table}

\section{Limitations}
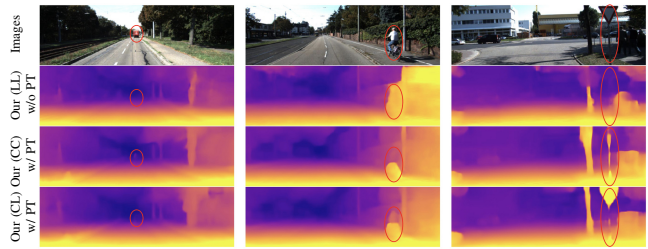
\begin{figure}[htp]
    \centering
    \begin{overpic}[scale=0.080]{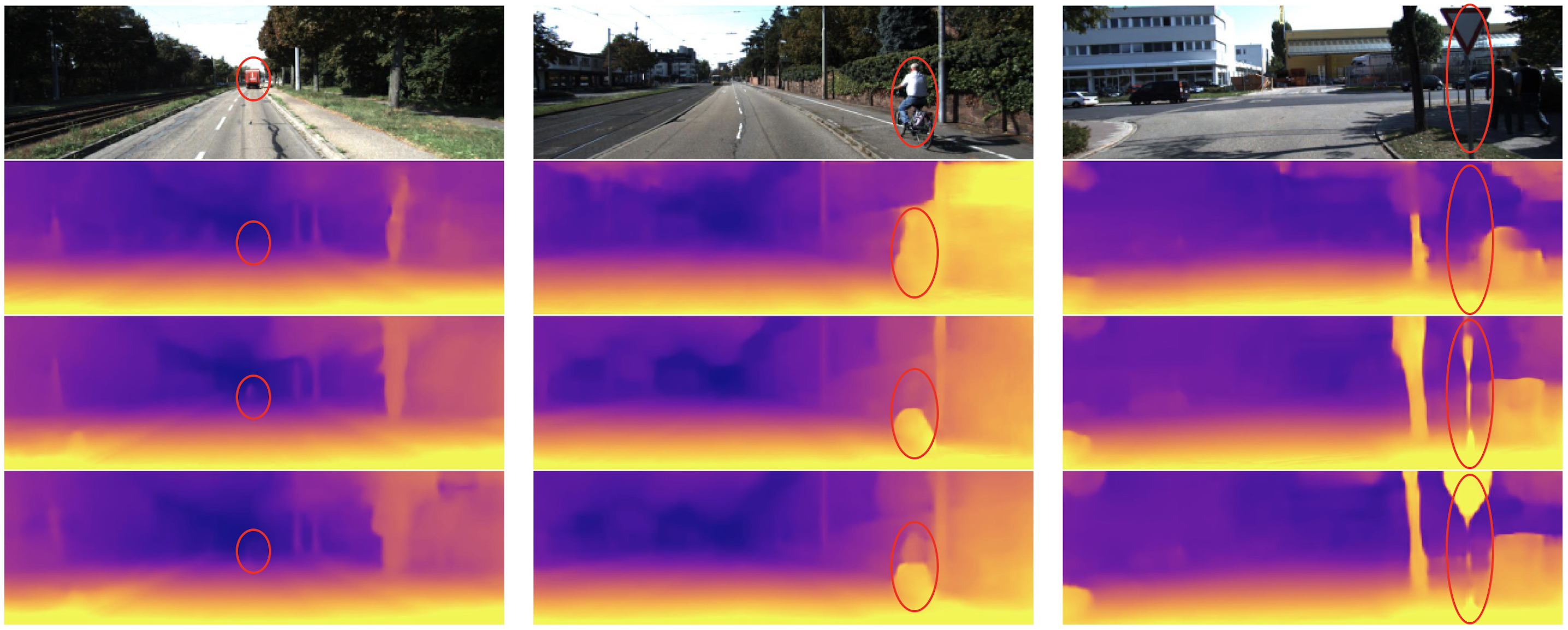}
    \put(-4.5,32){\tiny\rotatebox{90}{Images}}
    \put(-4.5,21){\tiny\rotatebox{90}{Our (LL)}}  
    \put(-2,22){\tiny\rotatebox{90}{w/o PT}} 
    \put(-4.5,10){\tiny\rotatebox{90}{Our (CC)}}  
    \put(-2,12.6){\tiny\rotatebox{90}{w/ PT}} 
    \put(-4.5,1){\tiny\rotatebox{90}{Our (CL)}}  
    \put(-2,2.5){\tiny\rotatebox{90}{w/ PT}} 
    \end{overpic}
    \caption{Failure cases of the proposed model. Left to right (columns): far objects, moving objects, and slender objects. CC, CL, and LL represent the different network architectures, in which C and L represent convolutional neural networks (CNNs) and the proposed encoder/decoder. For instance, CL indicates that the encoder and decoder adopt CNNs and the proposed decoder, respectively. PT indicates pretraining.}
    \label{failurecases}
\end{figure}

In this section, some failure cases of the proposed model are discussed. As shown in Fig. \ref{failurecases}, we find that it is challenging to 1) capture extremely far objects, which requires more accurate pose estimation; 2) accurately capture moving objects because such objects violate the static scene assumption; and 3) predict slender objects because such objects are easily confused as being part of the background. We will work on these issues in future studies.

\section{Conclusion}
This paper focuses on how to simultaneously extract global and local information from images and predict a geometrically smooth depth map. For extraction, an innovative network, DLNet, is proposed along with special depth and pose decoders. For prediction, a 3DGS is presented. Moreover, we explore the multi-scale prediction strategy used for overcoming the gradient locality issue and propose a maximum margin dual-scale prediction (MMDSP) strategy for efficient and effective predictions. Detailed experiments demonstrate that the proposed model achieves competitive performance against state-of-the-art methods and reduces time and space complexities by more than  $62\%$ and $56\%$, respectively. We hope that our work contributes to the relevant academic and industrial communities.

\ifCLASSOPTIONcaptionsoff
  \newpage
\fi



%

\bibliography{IEEEabrv,DepthLinformer.bib}{}

\begin{thebibliography}{10}
\providecommand{\url}[1]{#1}
\csname url@samestyle\endcsname
\providecommand{\newblock}{\relax}
\providecommand{\bibinfo}[2]{#2}
\providecommand{\BIBentrySTDinterwordspacing}{\spaceskip=0pt\relax}
\providecommand{\BIBentryALTinterwordstretchfactor}{4}
\providecommand{\BIBentryALTinterwordspacing}{\spaceskip=\fontdimen2\font plus
\BIBentryALTinterwordstretchfactor\fontdimen3\font minus
  \fontdimen4\font\relax}
\providecommand{\BIBforeignlanguage}[2]{{%
\expandafter\ifx\csname l@#1\endcsname\relax
\typeout{** WARNING: IEEEtran.bst: No hyphenation pattern has been}%
\typeout{** loaded for the language `#1'. Using the pattern for}%
\typeout{** the default language instead.}%
\else
\language=\csname l@#1\endcsname
\fi
#2}}
\providecommand{\BIBdecl}{\relax}
\BIBdecl

\bibitem{godard2019digging}
C.~Godard, O.~Mac~Aodha, M.~Firman, and G.~J. Brostow, ``Digging into
  self-supervised monocular depth estimation,'' in \emph{Proc. IEEE Int. Conf.
  Comput. Vision}, 2019, pp. 3828--3838.

\bibitem{bian2019unsupervised}
J.~Bian, Z.~Li, N.~Wang, H.~Zhan, C.~Shen, M.-M. Cheng, and I.~Reid,
  ``Unsupervised scale-consistent depth and ego-motion learning from monocular
  video,'' in \emph{Adv. Neur. In. (NeurIPS)}, 2019, pp. 35--45.

\bibitem{yang2017unsupervised}
Z.~Yang, P.~Wang, W.~Xu, L.~Zhao, and R.~Nevatia, ``Unsupervised learning of
  geometry with edge-aware depth-normal consistency,'' \emph{arXiv preprint
  arXiv:1711.03665}, 2017.

\bibitem{zhou2017unsupervised}
T.~Zhou, M.~Brown, N.~Snavely, and D.~G. Lowe, ``Unsupervised learning of depth
  and ego-motion from video,'' in \emph{Proc. IEEE Conf. Comput. Vis. Pattern
  Recognit. (CVPR)}, 2017, pp. 1851--1858.

\bibitem{yin2018geonet}
Z.~Yin and J.~Shi, ``Geonet: Unsupervised learning of dense depth, optical flow
  and camera pose,'' in \emph{Proc. IEEE Conf. Comput. Vis. Pattern Recognit.
  (CVPR)}, 2018, pp. 1983--1992.

\bibitem{geiger2013vision}
A.~Geiger, P.~Lenz, C.~Stiller, and R.~Urtasun, ``Vision meets robotics: The
  kitti dataset,'' \emph{Int. J. Rob. Res.}, vol.~32, no.~11, pp. 1231--1237,
  2013.

\bibitem{absolutedepthXue2020}
F.~Xue, G.~Zhuo, Z.~Huang, W.~Fu, Z.~Wu, and M.~H.~A. Jr, ``Toward hierarchical
  self-supervised monocular absolute depth estimation for autonomous driving
  applications,'' \emph{arXiv preprint arXiv:2004.05560}, 2020.

\bibitem{cao2017estimating}
Y.~Cao, Z.~Wu, and C.~Shen, ``Estimating depth from monocular images as
  classification using deep fully convolutional residual networks,'' \emph{IEEE
  Trans. Circuits Syst. Video Technol.}, vol.~28, no.~11, pp. 3174--3182, 2017.

\bibitem{eigen2015predicting}
D.~Eigen and R.~Fergus, ``Predicting depth, surface normals and semantic labels
  with a common multi-scale convolutional architecture,'' in \emph{Proc. IEEE
  Int. Conf. Comput. Vision}, 2015, pp. 2650--2658.

\bibitem{li2015depth}
B.~Li, C.~Shen, Y.~Dai, A.~Van Den~Hengel, and M.~He, ``Depth and surface
  normal estimation from monocular images using regression on deep features and
  hierarchical crfs,'' in \emph{Proc. IEEE Conf. Comput. Vis. Pattern Recognit.
  (CVPR)}, 2015, pp. 1119--1127.

\bibitem{liu2015deep}
F.~Liu, C.~Shen, and G.~Lin, ``Deep convolutional neural fields for depth
  estimation from a single image,'' in \emph{Proc. IEEE Conf. Comput. Vis.
  Pattern Recognit. (CVPR)}, 2015, pp. 5162--5170.

\bibitem{mousavian2016joint}
A.~Mousavian, H.~Pirsiavash, and J.~Ko{\v{s}}eck{\'a}, ``Joint semantic
  segmentation and depth estimation with deep convolutional networks,'' in
  \emph{Proc. - Int. Conf. 3D Vis. (3DV)}.\hskip 1em plus 0.5em minus
  0.4em\relax IEEE, 2016, pp. 611--619.

\bibitem{xu2017multi}
D.~Xu, E.~Ricci, W.~Ouyang, X.~Wang, and N.~Sebe, ``Multi-scale continuous crfs
  as sequential deep networks for monocular depth estimation,'' in \emph{Proc.
  IEEE Conf. Comput. Vis. Pattern Recognit. (CVPR)}, 2017, pp. 5354--5362.

\bibitem{xu2018structured}
D.~Xu, W.~Wang, H.~Tang, H.~Liu, N.~Sebe, and E.~Ricci, ``Structured attention
  guided convolutional neural fields for monocular depth estimation,'' in
  \emph{Proc. IEEE Conf. Comput. Vis. Pattern Recognit. (CVPR)}, 2018, pp.
  3917--3925.

\bibitem{karschdepth}
K.~Karsch, C.~Liu, and S.~Kang, ``Depth extraction from video using
  non-parametric sampling-supplemental material,'' in \emph{Proc. Eur. Conf.
  Comput. Vis., vol. part V}.\hskip 1em plus 0.5em minus 0.4em\relax Citeseer,
  2012, pp. 775--788.

\bibitem{saxena20083}
A.~Saxena, S.~H. Chung, and A.~Y. Ng, ``3-d depth reconstruction from a single
  still image,'' \emph{Int. J. Comput. Vis.}, vol.~76, no.~1, pp. 53--69, 2008.

\bibitem{Linformer}
S.~Wang, B.~Z. Li, M.~Khabsa, H.~Fang, and H.~Ma, ``Linformer: Self-attention
  with linear complexity,'' \emph{arXiv preprint arXiv:2006.04768}, 2020.

\bibitem{Transformer}
A.~Vaswani, N.~Shazeer, N.~Parmar, J.~Uszkoreit, L.~Jones, A.~N. Gomez,
  L.~Kaiser, and I.~Polosukhin, ``Attention is all you need,'' \emph{arXiv
  preprint arXiv:1706.03762}, 2017.

\bibitem{saxena2008make3d}
A.~Saxena, M.~Sun, and A.~Y. Ng, ``Make3d: Learning 3d scene structure from a
  single still image,'' \emph{IEEE Trans. Pattern Anal. Mach. Intell.},
  vol.~31, no.~5, pp. 824--840, 2008.

\bibitem{cordts2016cityscapes}
M.~Cordts, M.~Omran, S.~Ramos, T.~Rehfeld, M.~Enzweiler, R.~Benenson,
  U.~Franke, S.~Roth, and B.~Schiele, ``The cityscapes dataset for semantic
  urban scene understanding,'' in \emph{Proceedings of the IEEE conference on
  computer vision and pattern recognition}, 2016, pp. 3213--3223.

\bibitem{baig2016coupled}
M.~H. Baig and L.~Torresani, ``Coupled depth learning,'' in \emph{IEEE Winter
  Conf. Appl. Comput. Vis. (WACV)}, 2016, pp. 1--10.

\bibitem{choi2015depth}
S.~Choi, D.~Min, B.~Ham, Y.~Kim, C.~Oh, and K.~Sohn, ``Depth analogy:
  Data-driven approach for single image depth estimation using gradient
  samples,'' \emph{IEEE Trans. Image Process.}, vol.~24, no.~12, pp.
  5953--5966, 2015.

\bibitem{furukawa2017depth}
R.~Furukawa, R.~Sagawa, and H.~Kawasaki, ``Depth estimation using structured
  light flow--analysis of projected pattern flow on an object's surface,'' in
  \emph{Proc. IEEE Int. Conf. Comput. Vision}, 2017, pp. 4640--4648.

\bibitem{zoran2015learning}
D.~Zoran, P.~Isola, D.~Krishnan, and W.~T. Freeman, ``Learning ordinal
  relationships for mid-level vision,'' in \emph{Proc. IEEE Int. Conf. Comput.
  Vision}, 2015, pp. 388--396.

\bibitem{chen2016single}
W.~Chen, Z.~Fu, D.~Yang, and J.~Deng, ``Single-image depth perception in the
  wild,'' in \emph{Adv. Neur. In. (NeurIPS)}, 2016, pp. 730--738.

\bibitem{eigen2014depth}
D.~Eigen, C.~Puhrsch, and R.~Fergus, ``Depth map prediction from a single image
  using a multi-scale deep network,'' in \emph{Adv. Neur. In. (NeurIPS)}, 2014,
  pp. 2366--2374.

\bibitem{laina2016deeper}
I.~Laina, C.~Rupprecht, V.~Belagiannis, F.~Tombari, and N.~Navab, ``Deeper
  depth prediction with fully convolutional residual networks,'' in \emph{Proc.
  - Int. Conf. 3D Vis. (3DV)}.\hskip 1em plus 0.5em minus 0.4em\relax IEEE,
  2016, pp. 239--248.

\bibitem{li2017two}
J.~Li, R.~Klein, and A.~Yao, ``A two-streamed network for estimating
  fine-scaled depth maps from single rgb images,'' in \emph{Proc. IEEE Int.
  Conf. Comput. Vision}, 2017, pp. 3372--3380.

\bibitem{almalioglu2019ganvo}
Y.~Almalioglu, M.~R.~U. Saputra, P.~P. de~Gusmao, A.~Markham, and N.~Trigoni,
  ``Ganvo: Unsupervised deep monocular visual odometry and depth estimation
  with generative adversarial networks,'' in \emph{Proc. IEEE Int. Conf. Rob.
  Autom.}, 2019, pp. 5474--5480.

\bibitem{cs2018depthnet}
A.~CS~Kumar, S.~M. Bhandarkar, and M.~Prasad, ``Depthnet: A recurrent neural
  network architecture for monocular depth prediction,'' in \emph{Proc. IEEE
  Conf. Comput. Vis. Pattern Recognit. (CVPR)}, 2018, pp. 283--291.

\bibitem{grigorev2017depth}
A.~Grigorev, F.~Jiang, S.~Rho, W.~J. Sori, S.~Liu, and S.~Sai, ``Depth
  estimation from single monocular images using deep hybrid network,''
  \emph{Multimed. Tools Appl.}, vol.~76, no.~18, pp. 18\,585--18\,604, 2017.

\bibitem{mancini2017toward}
M.~Mancini, G.~Costante, P.~Valigi, T.~A. Ciarfuglia, J.~Delmerico, and
  D.~Scaramuzza, ``Toward domain independence for learning-based monocular
  depth estimation,'' \emph{IEEE Robot. Autom. Let.}, vol.~2, no.~3, pp.
  1778--1785, 2017.

\bibitem{tananaev2018temporally}
D.~Tananaev, H.~Zhou, B.~Ummenhofer, and T.~Brox, ``Temporally consistent depth
  estimation in videos with recurrent architectures,'' in \emph{Proc. Eur.
  Conf. Comput. Vis.}, 2018, pp. 0--0.

\bibitem{wang2019recurrent}
R.~Wang, S.~M. Pizer, and J.-M. Frahm, ``Recurrent neural network for (un-)
  supervised learning of monocular video visual odometry and depth,'' in
  \emph{Proc. IEEE Conf. Comput. Vis. Pattern Recognit. (CVPR)}, 2019, pp.
  5555--5564.

\bibitem{mypaper2020}
S.~Jia, X.~Pei, Z.~Yang, S.~Tian, and Y.~Yue, ``Novel hybrid neural network for
  dense depth estimation using on-board monocular images,''
  \emph{Transportation research record}, vol. 2674, no.~12, pp. 312--323, 2020.

\bibitem{fu2018deep}
H.~Fu, M.~Gong, C.~Wang, K.~Batmanghelich, and D.~Tao, ``Deep ordinal
  regression network for monocular depth estimation,'' in \emph{Proc. IEEE
  Conf. Comput. Vis. Pattern Recognit. (CVPR)}, 2018, pp. 2002--2011.

\bibitem{chen2019towards}
P.-Y. Chen, A.~H. Liu, Y.-C. Liu, and Y.-C.~F. Wang, ``Towards scene
  understanding: Unsupervised monocular depth estimation with semantic-aware
  representation,'' in \emph{Proc. IEEE Conf. Comput. Vis. Pattern Recognit.
  (CVPR)}, 2019, pp. 2624--2632.

\bibitem{garg2016unsupervised}
R.~Garg, V.~K. Bg, G.~Carneiro, and I.~Reid, ``Unsupervised cnn for single view
  depth estimation: Geometry to the rescue,'' in \emph{Lect. Notes Comput.
  Sci.}\hskip 1em plus 0.5em minus 0.4em\relax Springer, 2016, pp. 740--756.

\bibitem{ranjan2019competitive}
A.~Ranjan, V.~Jampani, L.~Balles, K.~Kim, D.~Sun, J.~Wulff, and M.~J. Black,
  ``Competitive collaboration: Joint unsupervised learning of depth, camera
  motion, optical flow and motion segmentation,'' in \emph{Proc. IEEE Conf.
  Comput. Vis. Pattern Recognit. (CVPR)}, 2019, pp. 12\,240--12\,249.

\bibitem{zhan2018unsupervised}
H.~Zhan, R.~Garg, C.~Saroj~Weerasekera, K.~Li, H.~Agarwal, and I.~Reid,
  ``Unsupervised learning of monocular depth estimation and visual odometry
  with deep feature reconstruction,'' in \emph{Proc. IEEE Conf. Comput. Vis.
  Pattern Recognit. (CVPR)}, 2018, pp. 340--349.

\bibitem{zhou2019unsupervised}
J.~Zhou, Y.~Wang, K.~Qin, and W.~Zeng, ``Unsupervised high-resolution depth
  learning from videos with dual networks,'' in \emph{Proc. IEEE Int. Conf.
  Comput. Vision}, 2019, pp. 6872--6881.

\bibitem{godard2017unsupervised}
C.~Godard, O.~Mac~Aodha, and G.~J. Brostow, ``Unsupervised monocular depth
  estimation with left-right consistency,'' in \emph{Proc. IEEE Conf. Comput.
  Vis. Pattern Recognit. (CVPR)}, 2017, pp. 270--279.

\bibitem{kuznietsov2017semi}
Y.~Kuznietsov, J.~Stuckler, and B.~Leibe, ``Semi-supervised deep learning for
  monocular depth map prediction,'' in \emph{Proc. IEEE Conf. Comput. Vis.
  Pattern Recognit. (CVPR)}, 2017, pp. 6647--6655.

\bibitem{feng2019sganvo}
T.~Feng and D.~Gu, ``Sganvo: Unsupervised deep visual odometry and depth
  estimation with stacked generative adversarial networks,'' \emph{IEEE Robot.
  Autom. Let.}, vol.~4, no.~4, pp. 4431--4437, 2019.

\bibitem{casser2019depth}
V.~Casser, S.~Pirk, R.~Mahjourian, and A.~Angelova, ``Depth prediction without
  the sensors: Leveraging structure for unsupervised learning from monocular
  videos,'' in \emph{Proceedings of the AAAI Conference on Artificial
  Intelligence}, vol.~33, 2019, pp. 8001--8008.

\bibitem{mypaper2021}
S.~Jia, X.~Pei, X.~Jing, and D.~Yao, ``Self-supervised 3d reconstruction and
  ego-motion estimation via on-board monocular video,'' \emph{IEEE Transactions
  on Intelligent Transportation Systems}, pp. 1--13, 2021.

\bibitem{park2019high}
K.~Park, S.~Kim, and K.~Sohn, ``High-precision depth estimation using
  uncalibrated lidar and stereo fusion,'' \emph{IEEE Trans. Intell. Transp.
  Syst.}, vol.~21, no.~1, pp. 321--335, 2019.

\bibitem{yang2019fast}
X.~Yang, J.~Chen, Y.~Dang, H.~Luo, Y.~Tang, C.~Liao, P.~Chen, and K.-T. Cheng,
  ``Fast depth prediction and obstacle avoidance on a monocular drone using
  probabilistic convolutional neural network,'' \emph{IEEE Trans. Intell.
  Transp. Syst.}, 2019.

\bibitem{T4C1}
N.~Carion, F.~Massa, G.~Synnaeve, N.~Usunier, A.~Kirillov, and S.~Zagoruyko,
  ``End-to-end object detection with transformers. in european conference on
  computer vision,'' in \emph{Proceedings of the European Conference on
  Computer Vision (ECCV)}.\hskip 1em plus 0.5em minus 0.4em\relax Springer,
  2020, pp. 213--229.

\bibitem{T4C2}
X.~Zhu, W.~Su, L.~Lu, B.~Li, X.~Wang, and J.~Dai, ``Deformable detr: Deformable
  transformers for end-to-end object detection,'' \emph{arXiv preprint
  arXiv:2010.04159}, 2020.

\bibitem{T4C3}
Y.~Wang, Z.~Xu, X.~Wang, C.~Shen, B.~Cheng, H.~Shen, and H.~Xia, ``End-to-end
  video instance segmentation with transformers,'' \emph{arXiv preprint
  arXiv:2011.14503}, 2020.

\bibitem{T4C4}
H.~Chen, Y.~Wang, T.~Guo, C.~Xu, Y.~Deng, Z.~Liu, S.~Ma, C.~Xu, and W.~Gao,
  ``Pre-trained image processing transformer,'' \emph{arXiv preprint
  arXiv:2012.00364}, 2020.

\bibitem{C4T1}
A.~Dosovitskiy, L.~Beyer, A.~Kolesnikov, D.~Weissenborn, X.~Zhai,
  T.~Unterthiner, M.~Dehghani, M.~Minderer, G.~Heigold, S.~Gelly, J.~Uszkoreit,
  and N.~Houlsby, ``An image is worth 16x16 words: Transformers for image
  recognition at scale,'' \emph{arXiv preprint arXiv:2010.11929}, 2020.

\bibitem{C4T2}
H.~Touvron, M.~Cord, M.~Douze, F.~Massa, A.~Sablayrolles, and H.~Jegou,
  ``Training data-efficient image transformers \& distillation through
  attention,'' \emph{arXiv preprint arXiv:2012.12877}, 2020.

\bibitem{C4T3}
S.~Zheng, J.~Lu, H.~Zhao, X.~Zhu, Z.~Luo, Y.~Wang, Y.~Fu, J.~Feng, T.~Xiang,
  P.~H. Torr, and L.~Zhang, ``Rethinking semantic segmentation from a
  sequence-to-sequence perspective with transformers,'' \emph{arXiv preprint
  arXiv:2012.15840}, 2020.

\bibitem{C4T4}
L.~Yuan, Y.~Chen, T.~Wang, W.~Yu, Y.~Shi, F.~E. Tay, J.~Feng, and S.~Yan,
  ``Tokens-to-token vit: Training vision transformers from scratch on
  imagenet,'' \emph{arXiv preprint arXiv:2101.11986}, 2021.

\bibitem{basri2003lambertian}
R.~Basri and D.~W. Jacobs, ``Lambertian reflectance and linear subspaces,''
  \emph{IEEE Trans. Pattern Anal. Mach. Intell.}, vol.~25, no.~2, pp. 218--233,
  2003.

\bibitem{wang2018learning}
C.~Wang, J.~Miguel~Buenaposada, R.~Zhu, and S.~Lucey, ``Learning depth from
  monocular videos using direct methods,'' in \emph{Proc. IEEE Conf. Comput.
  Vis. Pattern Recognit. (CVPR)}, 2018, pp. 2022--2030.

\bibitem{liu2015learning}
F.~Liu, C.~Shen, G.~Lin, and I.~Reid, ``Learning depth from single monocular
  images using deep convolutional neural fields,'' \emph{IEEE Trans. Pattern
  Anal. Mach. Intell.}, vol.~38, no.~10, pp. 2024--2039, 2015.

\bibitem{karsch2014depth}
K.~Karsch, C.~Liu, and S.~B. Kang, ``Depth transfer: Depth extraction from
  video using non-parametric sampling,'' \emph{IEEE Trans. Pattern Anal. Mach.
  Intell.}, vol.~36, no.~11, pp. 2144--2158, 2014.

\bibitem{liu2014discrete}
M.~Liu, M.~Salzmann, and X.~He, ``Discrete-continuous depth estimation from a
  single image,'' in \emph{Proc. IEEE Conf. Comput. Vis. Pattern Recognit.
  (CVPR)}, 2014, pp. 716--723.

\bibitem{kingma2014adam}
D.~P. Kingma and J.~Ba, ``Adam: A method for stochastic optimization,''
  \emph{arXiv preprint arXiv:1412.6980}, 2014.

\bibitem{relu}
V.~Nair and G.~E. Hinton, ``Rectified linear units improve restricted boltzmann
  machines,'' in \emph{Icml}, 2010.

\bibitem{elu}
D.-A. Clevert, T.~Unterthiner, and S.~Hochreiter, ``Fast and accurate deep
  network learning by exponential linear units (elus),'' \emph{arXiv preprint
  arXiv:1511.07289}, 2015.

\bibitem{gelu}
D.~Hendrycks and K.~Gimpel, ``Gaussian error linear units (gelus),''
  \emph{arXiv preprint arXiv:1606.08415}, 2016.

\end{thebibliography}
\bibliographystyle{IEEEtran}

\end{document}